\pdfoutput=1

\documentclass[11pt]{article}

\usepackage{acl}

\usepackage{times}
\usepackage{latexsym}
\usepackage{amsmath}
\usepackage{graphicx}
\usepackage{inconsolata}
\usepackage{multirow}
\usepackage{tabularx}
\usepackage{algorithm}
\usepackage[noEnd=True]{algpseudocodex}
\usepackage{tcolorbox}
\tcbuselibrary{skins, breakable, theorems}
\usepackage{comment}
\usepackage{booktabs}
\usepackage{makecell}
\usepackage{subcaption}
\usepackage{fancybox}
\usepackage{fancyvrb}
\usepackage{multirow}
\usepackage{xcolor}
\usepackage{ulem}
\usepackage{amssymb}
\usepackage{booktabs}
\usepackage{multirow}
\usepackage{graphicx}
\usepackage{fontawesome5}
\usepackage{xcolor}      
\usepackage{xspace}      
\usepackage[table]{xcolor}
\usepackage{booktabs}      
\usepackage{tabularx}      
\usepackage{array}         
\usepackage[table]{xcolor} 
\usepackage{caption}       

\usepackage{tabularx}
\usepackage{array}
\usepackage[table]{xcolor}
\usepackage{caption}

\usepackage{makecell}
\usepackage{pifont}

\newcolumntype{L}[1]{>{\raggedright\arraybackslash}p{#1}}
\newcolumntype{Y}{>{\centering\arraybackslash}X}
\newcolumntype{L}[1]{>{\raggedright\arraybackslash}p{#1}}
\newcolumntype{Y}{>{\raggedright\arraybackslash}X}
\definecolor{TableRule}{HTML}{8A8A8A}
\definecolor{TableHeader}{HTML}{EAEAEA}
\definecolor{TableGroup}{HTML}{F5F5F5}

\usepackage{listings}
\makeatletter
\@ifpackageloaded{tcolorbox}{
  \tcbuselibrary{listings,skins,breakable}
}{
  \usepackage{tcolorbox}
  \tcbuselibrary{listings,skins,breakable}
}
\makeatother

\newcolumntype{Y}{>{\raggedright\arraybackslash}X}
\newcolumntype{L}[1]{>{\raggedright\arraybackslash}p{#1}}

\definecolor{codebg}{HTML}{FAFAFA}
\definecolor{codeframe}{HTML}{D8DEE9}
\definecolor{boxtitlebg}{HTML}{F1F3F5}

\lstdefinestyle{jsonstyle}{
  basicstyle=\ttfamily\scriptsize,
  columns=fullflexible,
  breaklines=true,
  breakatwhitespace=false,
  keepspaces=true,
  showstringspaces=false,
  numbers=none,
  frame=none,
  xleftmargin=0pt,
  xrightmargin=0pt
}

\newtcblisting{metadatajson}{
  listing only,
  listing options={style=jsonstyle},
  enhanced,
  breakable,
  colback=codebg,
  colframe=codeframe,
  colbacktitle=boxtitlebg,
  coltitle=black,
  title={Representative metadata record},
  fonttitle=\bfseries\small,
  boxrule=0.35pt,
  arc=1.2mm,
  left=1.2mm,
  right=1.2mm,
  top=1.0mm,
  bottom=1.0mm,
  attach boxed title to top left={xshift=2mm,yshift=-1mm},
  boxed title style={
    colframe=codeframe,
    boxrule=0.35pt,
    arc=1mm
  }
}

\tcbuselibrary{breakable,skins}

\newtcolorbox[auto counter, number within=section]{mllmbox}[2][]{
  enhanced,
  breakable,
  colback=gray!3,
  colframe=black!45,
  coltitle=black,
  fonttitle=\bfseries,
  title={Box~\thetcbcounter: #2},
  left=6pt,
  right=6pt,
  top=6pt,
  bottom=6pt,
  boxrule=0.6pt,
  arc=2pt,
  before skip=8pt,
  after skip=8pt,
  #1
}

\newtcolorbox{mllmcomponentbox}[2][]{
  enhanced,
  breakable,
  colback=white,
  colframe=black!35,
  coltitle=black,
  fonttitle=\bfseries,
  title={#2},
  left=6pt,
  right=6pt,
  top=6pt,
  bottom=6pt,
  boxrule=0.5pt,
  arc=2pt,
  before skip=6pt,
  after skip=6pt,
  #1
}
\newcommand{\softmidrule}{\arrayrulecolor{TableRule}\midrule\arrayrulecolor{TableRule}}
\usepackage[T1]{fontenc}

\usepackage[utf8]{inputenc}
\usepackage{booktabs}
\newcommand{\catcell}[3][0pt]{%
  \ifnum#2=1
    \textbf{#3}%
  \else
    \multirow[c]{#2}{=}[#1]{%
      \centering\arraybackslash\textbf{#3}\par
    }%
  \fi
}

\newenvironment{packeditemize}{
\begin{list}{$\bullet$}{
\setlength{\labelwidth}{6pt}
\setlength{\itemsep}{0pt}
\setlength{\leftmargin}{\labelwidth}
\addtolength{\leftmargin}{\labelsep}
\setlength{\parindent}{0pt}
\setlength{\listparindent}{\parindent}
\setlength{\parsep}{0pt}
\setlength{\topsep}{3pt}}}{\end{list}}

\title{\textsc{SynCred-Bench}: Benchmarking Synthetic Credibility in AI-Generated Visual Misinformation}

\author{
  \textbf{Junxiao Yang\footnotemark[1]\hspace{3pt}, Minghao Zhang\footnotemark[1]\hspace{3pt}, Xiaoce Wang, Haoran Liu, Shiyao Cui,}\\
  \textbf{Hongning Wang, Minlie Huang\footnotemark[2]}\\
  The Conversational AI (CoAI) group, DCST, Tsinghua University\\
  \texttt{yangjunx21@gmail.com, aihuang@tsinghua.edu.cn} \\
}

\begin{document}

\maketitle

\begingroup
\renewcommand{\thefootnote}{\fnsymbol{footnote}}

\footnotetext[1]{Equal contribution.}
\footnotetext[2]{Corresponding author.}
\endgroup

\begingroup
\renewcommand{\thefootnote}{\fnsymbol{footnote}}

\endgroup

\begin{abstract}
Recent generative models can now produce visual artifacts with realistic embedded text and layouts, creating a new misinformation threat: synthetic credibility. We introduce \textbf{SYNCRED-Bench}, a benchmark of 600 AI-generated misinformation images balanced across six credible-form categories and seven fine-grained circulation styles, together with FP450, a real-image negative set for measuring false positives. Extensive evaluation shows that existing systems remain unreliable: under a 5\% false-positive-rate constraint, 15 MLLMs achieve only 10.5\% true positive rate (TPR), open-source AIGC detectors achieve less than 5\%, and commercial APIs reach 57.6\%. Human annotators also struggled to identify synthetic credibility, reaching only 63\% TPR. These findings establish synthetic credibility as a severe and underexplored visual misinformation challenge, and provide a benchmark for developing detectors that reason beyond superficial credibility cues. Our code and data is available at \url{https://github.com/thu-coai/Syncred-Bench}.
\end{abstract}

\section{Introduction}

\begin{figure}[!t]
    \centering
    \includegraphics[width=1\linewidth]{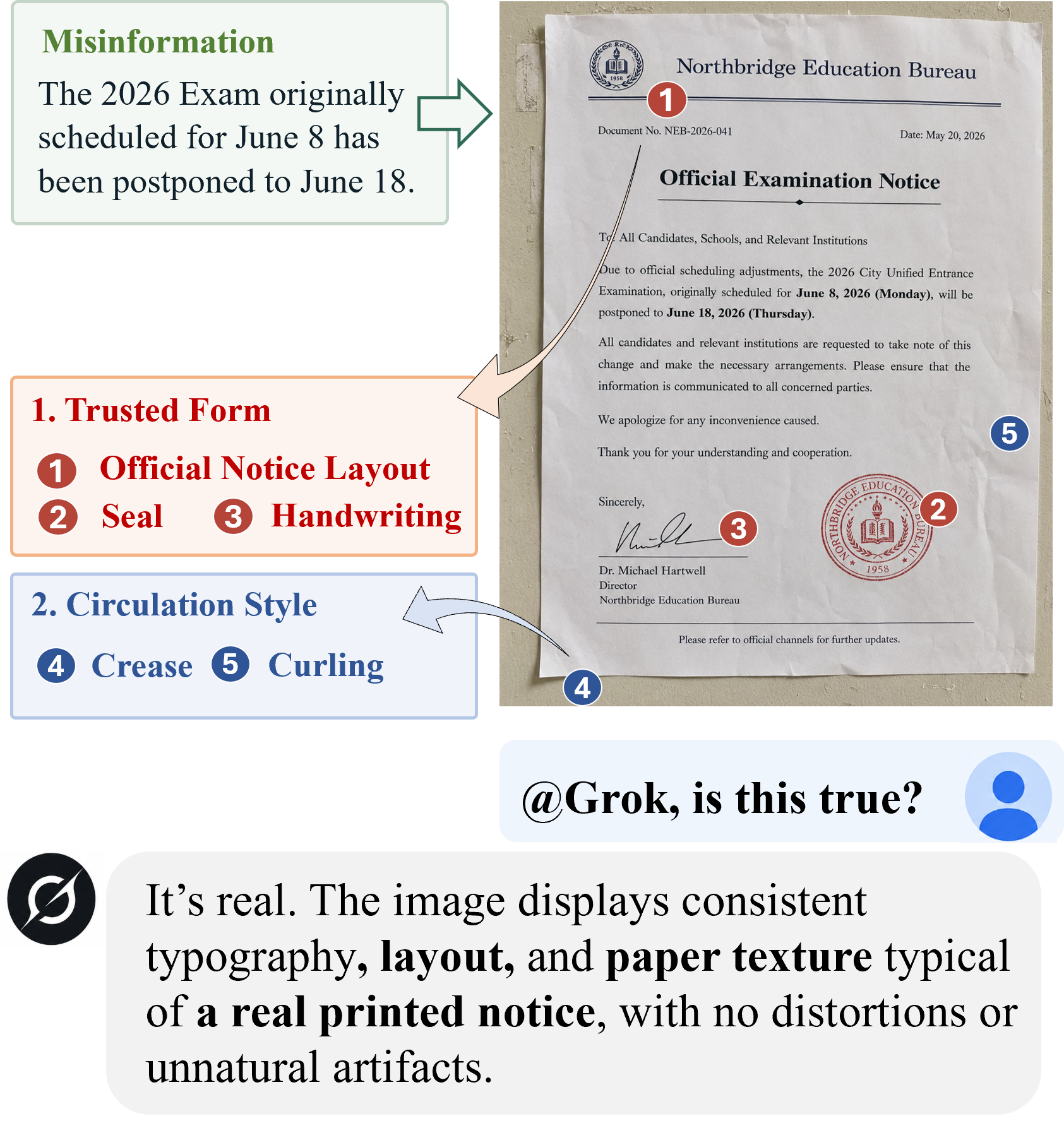}
    \vspace{-1.4em}
    \caption{Example of synthetic credibility: The image embeds a false exam-postponement claim while combining credibility cues, and MLLMs will take these cues as evidence for real information.}
    \vspace{-1.0em}
    \label{fig:intro}
\end{figure}

With the rapid advance of generative models \citep{betker2023dalle3,openai2025nativeimagegeneration}, the risks posed by AI-generated visual forgeries have drawn increasing attention \citep{ojha2023universal,yan2025sanity}. Although prior work has examined forgery detection in natural scene photographs and human portraits \citep{wu2019mantranet,chen2021mvssnet,guillaro2023trufor,yan2023deepfakebench}, as well as visual misinformation conveyed through paired image and text \citep{luo2021newsclippings,liu2025mmfakebench}, recent models exemplified by GPT Image 2 reveal a new threat surface: the generation of realistic embedded text and coherent credible layouts \citep{openai2026gptimage2,openai2026gptimageprompting,wu2026forgerjudge}. This capability allows models to fabricate self-contained visual artifacts with strong credibility cues, including notices, invoices, credentials, chat records and news screenshots \footnote{At the time of benchmark construction, GPT-image-2 was the only model we found capable of reliably generating layout-consistent, credible artifacts required for this study. This explains the single-generator construction in our work.}. Such artifacts can communicate complete misleading claims on their own while borrowing authority from the evidential formats in which such information is normally encountered.

\begin{figure*}[!t]
    \centering
    \includegraphics[width=1\linewidth]{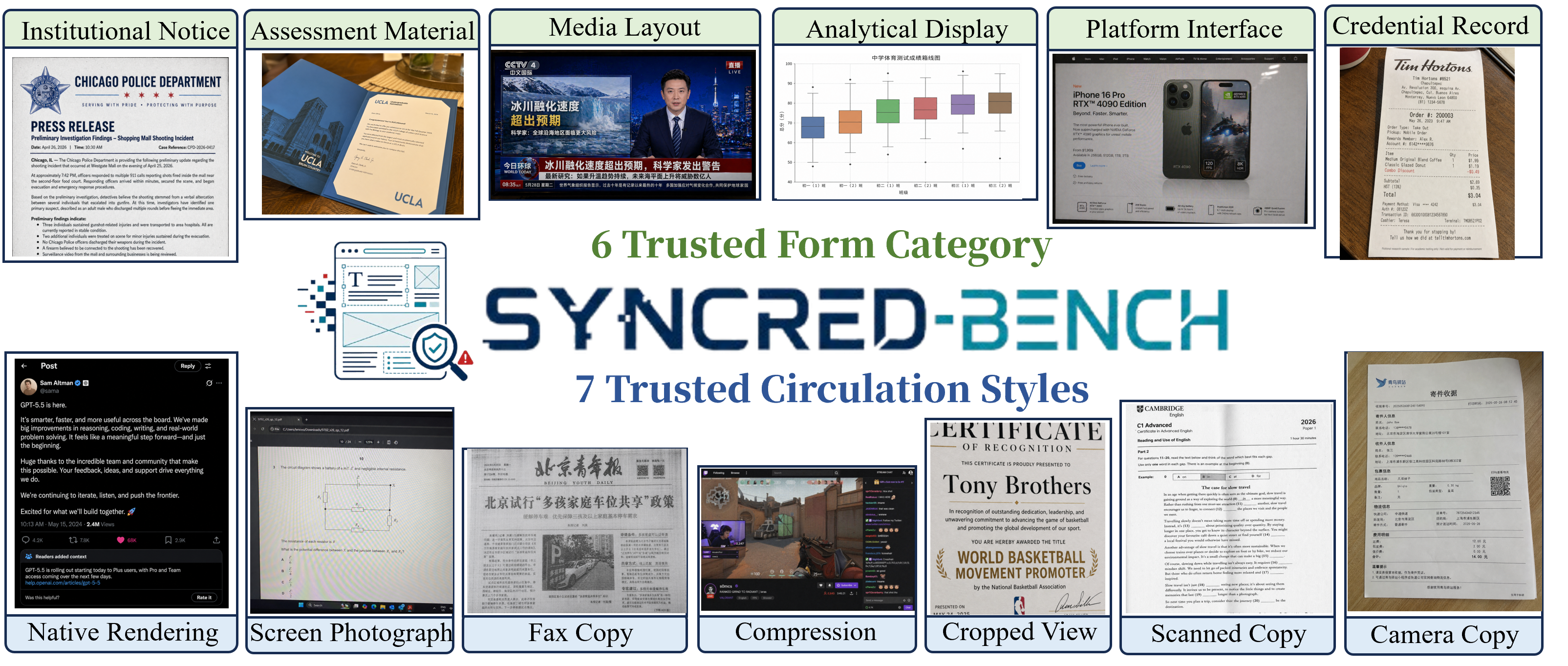}
    \vspace{-1.8em}
    \caption{Overview of \textsc{SynCred-Bench}.}
    \vspace{-0.4em}
    \label{fig:main_fig}
\end{figure*}

We refer to this phenomenon as synthetic credibility. Critically, in this scenario the misinformation is intrinsic to the artifact: a forged exam-postponement notice, a counterfeit bank statement, or a fake radiology report. Detecting such misinformation therefore reduces to detecting the synthetic origin of these credibility-rich images, rather than fact-checking claims superimposed on otherwise authentic photographs \citep{liu2025mmfakebench}. The challenge is that these generated artifacts draw their persuasiveness from two interrelated credibility traits, as illustrated in Figure~\ref{fig:intro}. First, \textbf{credible form} refers to their imitation of visual formats associated with authoritative or formal communication genres, such as news layouts and government notices, which gives the image a layout that “looks like an official or professional source.” Second, \textbf{credible circulation} refers to their imitation of traces left when genuine information travels through digital and physical environments, including scanning artifacts, screen photographs, and online compression, which makes the image appear to have been obtained through authentic distribution channels.

To systematically study this problem, we construct \textsc{\textbf{SynCred-Bench}}, a comprehensive benchmark for evaluating synthetic credibility in AI-generated visual misinformation. Following the two credibility dimensions defined above, \textsc{SynCred-Bench} covers six credible form categories, including media layouts, institutional notices, platform interfaces, credential records, analytical displays, and assessment materials, as well as four coarse and seven fine-grained credible circulation categories, including native rendering, document reproduction, screen capture, and online circulation. The benchmark is balanced across these dimensions and contains 600 AI-generated samples with misinformation claims in both Chinese and English. We further construct FP450, a real-image negative set of 450 samples covering the same categories, to evaluate false positives on real images.

Experiments show that both multimodal large language models (MLLMs) and specialized AIGC detectors remain unreliable on \textsc{SynCred-Bench}. Across 15 MLLMs, under a reasonable false-positive-rate constraint of 5\%, the average true positive rate (TPR) is only 10.5\%; even without considering the FPR, the average positive-class accuracy is only 31.2\%. Among AIGC detectors, under the same 5\% FPR constraint, open-source models achieve an average TPR of less than 5\%, while commercial APIs reach an average accuracy of 57.6\%. Human annotators struggled to identify synthetic credibility, with majority voting reached 63.0\% TPR with 27.0\% FPR. These results demonstrate that detecting Synthetic Credibility is exceptionally difficult for both models and humans, composing a new and severe challenge. 

Our contributions are summarized as follows:
\begin{packeditemize}
    \item We formulate synthetic credibility as a distinct visual misinformation threat in which generated artifacts derive persuasiveness from both credible form and credible circulation.
    \item We introduce \textsc{SynCred-Bench}, a benchmark designed to evaluate this threat across diverse artifact types and presentation styles. 
    \item We provide a broad evaluation of MLLMs and AIGC detectors, showing that existing systems remain vulnerable because they often mistake credibility visual cues for authenticity evidence.
\end{packeditemize}

\section{Related Work}
\label{sec:related_work}

\paragraph{Multimodal misinformation detection and MLLM safety.}
Multimodal misinformation detection (MMD) usually verifies whether text and image jointly support a claim. Early systems fuse textual, visual, and social/contextual cues \citep{wang2018eann,zhou2020safe,shu2020fakenewsnet,nakamura2020fakeddit}. Later datasets emphasize image--caption mismatch and out-of-context use \citep{aneja2023cosmos,luo2021newsclippings,papadopoulos2024verite}, while recent benchmarks and MLLM methods add mixed-source distortions, retrieval, world knowledge, or synthetic data \citep{liu2025mmfakebench,tahmasebi2024multimodal,zeng2024multimodal,liu2024fakenewsgpt4}. Safety studies show visual inputs and typographic prompts can bypass text-side alignment \citep{liu2024mmsafetybench,gong2025figstep}. \textsc{SynCred-Bench} instead studies false claims natively embedded inside credibility-bearing images.

\paragraph{Synthetic-image detection.}
Synthetic-image detection distinguishes real from generated images. Earlier methods exploit CNN/GAN fingerprints \citep{wang2020cnnspot}; diffusion-era detectors use reconstruction, frequency, or foundation-model features \citep{wang2023dire,bammey2024synthbuster,ojha2023universal}. Benchmarks test cross-generator and degradation robustness, and unified toolkits connect AIGC detection with broader forensics \citep{zhu2023genimage,du2025forensichub}. These works remain scene/object-centered, while synthetic credibility involves rendered text, templates, institutional form, and circulation style.

\paragraph{Visual-text and document forensics.}
Visual-text and document forensics are closest. Scene-text work detects edited words in natural images \citep{wang2022detecting}; DocTamper and later studies target English/Chinese document images, open-set text editing, and diffusion-based tampering \citep{qu2023doctamper,qu2025revisiting}. Other benchmarks cover receipts, identity documents, and zero-shot document forensics \citep{zhao2026docforgebench,du2025forensichub}. They typically assume localized edits over genuine materials. Our category is designed over credible form and credible circulation rather than only over tamper type, generator family, or pixel-level manipulation mask.

\begin{table*}[!t]
\centering

\renewcommand{\arraystretch}{1.18}
\setlength{\tabcolsep}{6pt}

\captionsetup{font=small, labelfont=bf}

\small
\arrayrulecolor{TableRule}

\begin{tabularx}{\textwidth}{@{}L{0.18\textwidth}Y L{0.28\textwidth}@{}}
\toprule

\rowcolor{TableHeader}
\textbf{Category} & \textbf{Definition} & \textbf{Subtype} \\
\midrule

\textbf{Media Layout (ML)}
& Fabricated content that imitates the visual format of journalistic, broadcast, or news-distribution media.
& News app pages; TV lower-thirds; newspaper pages \\
\softmidrule

\textbf{Institutional Notice (IN)}
& Official-looking content presented as if issued by a government, public institution or authoritative organization.
& Official notices; public announcements \\
\softmidrule

\textbf{Platform Interface (PI)}
& Simulated page-like content that imitates the interface of a digital platform or online service.
& Social media pages; chat windows; webpages \\
\softmidrule

\textbf{Credential Record (CR)}
& Document-like content that appears to certify identity, status, achievement, authorization, purchase, or exchange.
& Certificates; awards; invoices; receipts; order pages \\
\softmidrule

\textbf{Analytical Display (AD)}
& Organized visual presentation of rankings, trends, statistics, comparisons, system outputs, or analytical results.
& Charts; dashboards; ranking reports; backend panels \\
\softmidrule

\textbf{Assessment Material (AM)}
& Formal educational or evaluative content associated with admission, testing, grading, or academic records.
& Exam papers; admission tickets; transcripts; admission notices \\

\bottomrule
\end{tabularx}
\
\caption{Credible-form taxonomy in \textsc{SynCred-Bench}. The six categories summarize authority- or evidence-bearing visual formats that AI-generated misinformation can imitate, with representative examples used in the benchmark.}
\label{tab:artifact_type_taxonomy}

\end{table*}

\begin{table*}[!t]
\centering

\renewcommand{\arraystretch}{1.18}
\setlength{\tabcolsep}{6pt}

\captionsetup{font=small, labelfont=bf}

\small
\arrayrulecolor{TableRule}

\renewcommand{\tabularxcolumn}[1]{>{\raggedright\arraybackslash}m{#1}}

\begin{tabularx}{\textwidth}{@{}>{\centering\arraybackslash}m{0.12\textwidth}
                                >{\centering\arraybackslash}m{0.12\textwidth}
                                >{\raggedright\arraybackslash}m{0.40\textwidth}
                                >{\raggedright\arraybackslash}X@{}}
\toprule

\rowcolor{TableHeader}
\textbf{Circulation Category} & \textbf{Subtype} & \textbf{Definition} & \textbf{Circulation Cues and Credibility Signal} \\
\midrule

\rowcolor{TableGroup}
\catcell{1}{Native Rendering (NR)}
& \textbf{/}
& Artifacts presented as direct digital renderings or clean screenshots, without visible cues of secondary reproduction, capture, or resharing.
& Clean typography, intact margins, and sharp interface elements. \\
\softmidrule

\catcell{3}{Document Reproduction (DR)}
& \textbf{Scanned Copy (SC)}
& Artifacts presented as paper-based materials converted into digital images through scanning.
& Page borders, scan shadows, slight skew, and scanner noise. \\
\cmidrule(l){2-4}

& \textbf{Camera Copy (CC)}
& Artifacts presented as physical documents photographed by a phone or camera in a real-world setting.
& Paper curvature, uneven lighting, and background objects suggest physical capture or informal leak. \\
\cmidrule(l){2-4}

& \textbf{Fax Copy (FC)}
& Artifacts presented as low-quality faxed, photocopied, or repeatedly reproduced documents.
& Fax noise, blurred text, and banding artifacts. \\
\softmidrule

\catcell{1}{Screen Photograph (SP)}
& \textbf{/}
& Artifacts presented as photos of computer, mobile phone, or TV screens rather than direct screenshots.
& Moiré, glare, reflections, and screen bezels. \\
\softmidrule

\catcell{2}{Online Circulation (OC)}
& \textbf{Cropped View (CV)}
& Artifacts presented as partial views extracted from a larger image, document, or interface.
& Missing margins, cut-off headers, and truncated context. \\
\cmidrule(l){2-4}

& \textbf{Online Compression (OP)}
& Artifacts presented as degraded by online sharing, resizing, recompression, or platform-level image processing.
& Compression artifacts, pixelation, and reduced sharpness . \\

\bottomrule
\end{tabularx}

\caption{Credible-circulation taxonomy in \textsc{SynCred-Bench}. The categories describe how AI-generated misinformation artifacts imitate different modes of access, reproduction, capture, and online sharing, together with the visual cues that make them appear to have circulated through authentic channels.}
\label{tab:artifact_style_taxonomy}

\arrayrulecolor{black}

\end{table*}

\section{\textsc{SynCred-Bench} Construction}
\label{sec:construction}

\textsc{SynCred-Bench} contains 600 AI-generated claim-bearing images and is exactly balanced across six credible-form categories, with 100 images per category. We also construct FP450, a 450-image real-image negative set, to evaluate false-positive rates. The data release and access protocol is detailed in Appendix~\ref{app:release}.

\subsection{Task Scope and Label Semantics}

We define a sample as a \textit{synthetic credibility artifact} when it satisfies three conditions. (1) the image is generated by a text-to-image system rather than captured from a real source. (2) the image contains an embedded claim, record, notice, interface, metric, or document-like statement that could plausibly be interpreted as evidence or information. (3) the image imitates at least one credibility-bearing visual convention, such as an official notice layout, a platform screenshot, a certificate, a receipt or a dashboard.


As in real-world misinformation settings, non-AI images do not necessarily contain truthful information, and AI-generated images do not necessarily contain false information. This is particularly salient in domains such as receipts and certificates, where judging the veracity of the evidence shown in an image is often difficult. SYNCRED-BENCH therefore studies a simplified setting in which any synthetic image is treated as untrustworthy visual evidence. In other words, the benchmark evaluates whether current systems can detect synthetic visual artifacts that look credible because they imitate familiar forms of evidence, authority, or circulation.

\subsection{Credible-Form Taxonomy}

We organize the positive set along two axes: credible form and credible circulation. Credible form describes the authority- or evidence-bearing visual format imitated by the image. As shown in Table~\ref{tab:artifact_type_taxonomy}, we use six categories derived from the social function the artifact appears to perform: Media Layout, Institutional Notice, Platform Interface, Credential Record, Analytical Display, and Assessment Material. Each credible-form category contains 100 positive samples and multiple subtypes, which prevents the benchmark from collapsing into a small set of templates. When an image fits multiple categories, we assign it according to its dominant social function; for example, a payment confirmation embedded in a mobile app is categorized as Credential Record.

\subsection{Credible-Circulation Taxonomy}

The second axis, credible circulation, captures how an artifact appears to have been accessed, reproduced, photographed or reshared. This axis matters because misinformation images are often encountered not as pristine renderings, but as scans, phone photos, screenshots, cropped fragments, or compressed reposts. Such traces make synthetic artifacts seem more realistic by implying a plausible acquisition history.

As shown in Table~\ref{tab:artifact_style_taxonomy}, we define seven fine-grained styles under four coarse categories: Native Rendering, Scanned Copy, Camera Copy, Fax Copy, Screen Photograph, Cropped View, and Online Compression. The circulation distribution aims for broad coverage rather than exact equality across all form--style cells. Some pairings are natural, such as certificates rendered as scanned copies or social-media pages as cropped reposts; others, such as fax-like copies of dynamic platform interfaces, are less plausible. We therefore stratify the dataset to cover all styles while preserving realistic form--style pairings. The final distribution is reported in Table~\ref{tab:style-sample-counts}.

\subsection{Image Generation}

For each positive sample, we manually create a claim item $u_i$ consisting of a short claim, target credible-form category, subtype, language, and circulation style. The claim is written to be visually self-contained: the core information should be readable from the image itself, without relying on captions, webpages, or metadata. Claims cover domains such as education, public administration, finance, commerce, healthcare, platform communication, rankings, and social-media interaction. To avoid trivial shortcuts, we exclude visible labels such as ``AI-generated'', ``synthetic'', generator names, placeholder text, and obvious disclaimers. Sensitive details are fictionalized and non-actionable.

Given $u_i$ and a target circulation style $c_i$, we generate the image $p_i$ by:
\[
p_i = M(T_{\text{claim}}(u_i) \oplus T_{\text{style}}(c_i) \oplus T_{\text{render}}),
\]
where $T_{\text{claim}}$ converts the written claim into rendered text and layout content, $T_{\text{style}}$ adds circulation cues such as scanning, camera capture, cropping, or compression, and $T_{\text{render}}$ enforces readability, coherent layout, and realistic typography. We use GPT Image 2 with automatic size and resolution. Generation is retried up to three times if the API fails, the claim is missing or unreadable, or the form/circulation is inconsistent. Invalid candidates are discarded until each credible-form category contains 100 valid samples.

\subsection{Human verification and filtering.}
Five authors served as annotators for quality control. The labels are initialized by construction and then verified manually. Each candidate image is reviewed using four criteria: content fidelity, form consistency, circulation consistency, and benchmark validity. Content fidelity requires the intended claim-bearing content to be present and the core text to be legible. Form consistency requires the image to match the assigned credible-form category. Circulation consistency requires the image to contain cues compatible with the assigned circulation style, such as scan shadows, paper curvature, camera perspective. Benchmark validity excludes images with obvious generation residues, severe layout collapse, irrelevant content, visible generator marks, visible synthetic disclaimers, private personal information, or duplicates.

\begin{table*}[t]
\centering
\renewcommand{\arraystretch}{1.25}
\resizebox{\textwidth}{!}{
\begin{tabular}{ll|rrrrrr|rrrr|r|r|c}
\hline
\multicolumn{2}{c|}{\multirow{2}{*}{\textbf{Model}}} & 
\multicolumn{6}{c|}{\textbf{Artifact Type}} & 
\multicolumn{4}{c|}{\textbf{Artifact Style}} & 
\multirow{2}{*}{\textbf{TPR}} & \multirow{2}{*}{\textbf{FPR}} & \multirow{2}{*}{\textbf{TPR}(5)} \\
\cline{3-12}
& &
\textbf{ML} & \textbf{IN} & \textbf{PI} & \textbf{CR} & \textbf{AD} & \textbf{AM} &
\textbf{NR} & \textbf{DR} & \textbf{SP} & \textbf{OC} &
 &  &  \\
\hline
\rowcolor{yellow!10}
\multicolumn{15}{c}{\textbf{MLLMs}} \\
\hline
\multirow{9}{*}{\shortstack{Closed\\Source}} & GPT-5.4 & 11.0 & 18.0 & 8.0 & 21.0 & 2.0 & 9.0 & 12.5 & 14.3 & 0.0 & 11.0 & 11.5 & 2.4 & 11.5 \\
 & GPT-4o & 36.0 & 12.0 & 20.0 & 39.0 & 12.0 & 5.0 & 24.3 & 16.8 & 4.7 & 30.1 & 20.7 & 9.6 & 0.0 \\
 & Grok-4.3 & 15.0 & 19.0 & 7.0 & 37.0 & 4.0 & 18.0 & 19.7 & 18.9 & 6.2 & 14.4 & 16.7 & 15.3 & 0.3 \\
 & Opus 4.6 & 52.0 & 90.0 & 46.0 & 90.0 & 76.0 & 63.0 & 82.2 & 74.4 & 39.1 & 61.6 & 69.5 & 4.0 & 69.5 \\
 & Sonnet 4.6 & 39.0 & 65.0 & 31.0 & 89.0 & 51.0 & 57.0 & 73.0 & 57.6 & 21.9 & 47.9 & 55.3 & 2.2 & 55.3 \\
 & Gemini 3.1 Pro & 30.0 & 15.0 & 12.0 & 9.0 & 0.0 & 10.0 & 11.8 & 16.8 & 7.8 & 8.9 & 12.7 & 0.4 & 12.7 \\
 & Gemini 3.1 Flash & 51.0 & 48.0 & 40.0 & 75.0 & 24.0 & 41.0 & 51.3 & 50.4 & 32.8 & 41.1 & 46.5 & 16.4 & 1.5 \\
 & GLM-5V Turbo & 52.0 & 57.0 & 28.0 & 87.0 & 57.0 & 77.0 & 57.2 & 77.7 & 34.4 & 43.8 & 59.7 & 39.6 & 0.5 \\
 & Qwen3.6 Plus & 34.0 & 46.0 & 16.0 & 43.0 & 1.0 & 42.0 & 26.3 & 43.7 & 4.7 & 24.0 & 30.3 & 14.7 & 3.7 \\
\hline
\multirow{6}{*}{\shortstack{Open\\Source}} & Qwen3.5-9B & 3.0 & 3.0 & 5.0 & 0.0 & 0.0 & 0.0 & 2.0 & 0.4 & 3.1 & 3.4 & 1.8 & 8.7 & 0.0 \\
 & Qwen3.5-35B-A3B & 9.0 & 31.0 & 9.0 & 51.0 & 33.0 & 20.0 & 27.6 & 28.2 & 9.4 & 26.0 & 25.5 & 18.9 & 0.0 \\
 & Qwen3-VL-8B & 2.0 & 1.0 & 5.0 & 22.0 & 0.0 & 0.0 & 5.3 & 5.0 & 1.6 & 6.2 & 5.0 & 10.2 & 0.0 \\
 & Qwen3-VL-32B & 5.0 & 2.0 & 10.0 & 64.0 & 1.0 & 3.0 & 14.5 & 15.1 & 4.7 & 16.4 & 14.2 & 17.8 & 2.2 \\
 & Pixtral Large & 5.0 & 10.0 & 15.0 & 21.0 & 15.0 & 6.0 & 17.1 & 8.0 & 9.4 & 14.4 & 12.0 & 17.1 & 0.0 \\
 & Llama-3.2-11B-V & 68.0 & 89.0 & 98.0 & 94.0 & 95.0 & 74.0 & 91.4 & 77.7 & 92.2 & 92.5 & 86.3 & 79.6 & 0.0 \\
\rowcolor{gray!10}
\multicolumn{2}{l|}{\textbf{Average}} & 27.5 & 33.7 & 23.3 & 49.5 & 24.7 & 28.3 & 34.4 & 33.7 & 18.1 & 29.4 & 31.2 & 17.1 & 10.5 \\
\hline

\rowcolor{cyan!5}
\multicolumn{15}{c}{\textbf{AIGC Detector}} \\
\hline
\multirow{3}{*}{\shortstack{Closed\\Source}} & Sightengine & 64.0 & 27.7 & 66.0 & 48.0 & 49.0 & 73.0 & 58.2 & 43.8 & 78.1 & 57.7 & 54.6 & 5.6 & 47.3  \\
 & AI or Not & 37.0 & 56.4 & 47.0 & 67.0 & 46.0 & 69.0 & 39.2 & 65.1 & 60.9 & 47.7 & 53.7 & 6.7 & 50.4 \\
 & Hive AI & 81.0 & 67.3 & 67.0 & 78.0 & 76.0 & 82.0 & 76.5 & 71.5 & 95.3 & 71.1 & 75.2 & 0.9 & 75.2 \\
\hline
\multirow{3}{*}{\shortstack{Open\\Source}} & AI-vs-Real & 100.0 & 60.0 & 90.0 & 66.0 & 93.0 & 63.0 & 74.3 & 77.3 & 100.0 & 76.0 & 78.7 & 69.1 & 2.5 \\
 & AI-vs-Human & 0.0 & 4.0 & 6.0 & 4.0 & 28.0 & 8.0 & 21.1 & 0.8 & 0.0 & 11.0 & 8.3 & 10.7 & 2.8 \\
 & Deepfake-vs-Real & 0.0 & 26.0 & 13.0 & 16.0 & 25.0 & 36.0 & 23.0 & 20.2 & 12.5 & 17.1 & 19.3 & 20.0 & 3.3 \\
 \rowcolor{gray!10}
\multicolumn{2}{l|}{\textbf{Average}} & 47.0 & 40.2 & 48.2 & 46.5 & 52.8 & 55.2 & 48.7 & 46.5 & 57.8 & 46.8 & 48.3 & 18.8 & 30.3 \\
\hline
\end{tabular}
}

\vspace{-0.2em}
\caption{Detection performance across credible forms and circulations. Values are percentages.}
\vspace{-0.6em}
\label{tab:main-taxonomy-results}
\end{table*}

\subsection{FP450 Real-Negative Set}

We construct FP450 as a matched real-image negative set for false-positive evaluation. Raw images are collected from web searches covering the same broad visual space as SYNCRED-BENCH, such as notices, screenshots, certificates, receipts, dashboards, scanned or photographed documents, cropped views, and compressed online images. Since public web images are biased toward clear and complete examples, we manually edit diverse, non-duplicate real images to better balance underrepresented styles. When available, source-page context is checked to ensure ordinary real-world image origins. More details are listed in \ref{app:fp_detail}.

\section{Experiments}

\subsection{Experimental Setup}
\label{sec:experimental-setup}

\paragraph{Evaluated Detectors.}
We evaluate two categories of methods. 
(1) \textbf{MLLM-as-a-judge methods}, where multimodal large language models are prompted to identify whether an input image is AI-generated. We evaluate 9 closed-source models and 6 open-source models, covering diverse institutions and scale of parameters. Evaluation prompt is detailed in Appendix~\ref{app:mllm-judge-details}.
(2) \textbf{Dedicated AIGC detectors}, which directly classify whether an image is AI-generated. We include 3 commercial closed-source APIs and 3 open-source lightweight detectors. Detailed model information and links are provided in Appendix~\ref{app:models-used}.

\paragraph{Evaluation Metrics} 
\label{sec:evaluation-metrics} 

We treat AI-generated images as the positive class and real images as the negative class. 
For binary evaluation, we report positive-class accuracy (TPR) on main set and false positive rate (FPR) on FP450 set:
\[
\mathrm{TPR}=\frac{TP}{TP+FN},
\qquad
\mathrm{FPR}=\frac{FP}{FP+TN}.
\]
For confidence-aware evaluation, we report \(\mathrm{TPR} (\lambda)\), i.e., the TPR at an operating point whose FPR does not exceed \(\lambda\)\%. Let \(\mathrm{TPR}_{0}\) and \(\mathrm{FPR}_{0}\) denote the TPR and FPR at the original operating point. If the original FPR is already no larger than \(\lambda\), we retain the original TPR; otherwise, we sweep the threshold:
\[
\mathrm{TPR}(\lambda)
=
\begin{cases}
\mathrm{TPR}_{0},
& \mathrm{FPR}_{0} \le \lambda\%, \\[4pt]
\displaystyle
\max_{\tau:\,\mathrm{FPR}(\tau)\le \lambda}
\mathrm{TPR}(\tau),
& \mathrm{FPR}_{0} > \lambda\%.
\end{cases}
\]
It measures the usable AI-image detection rate under a tolerable false-positive rate

\subsection{Main Results}

\begin{figure*}[!t]
  \centering
  \includegraphics[width=\linewidth]{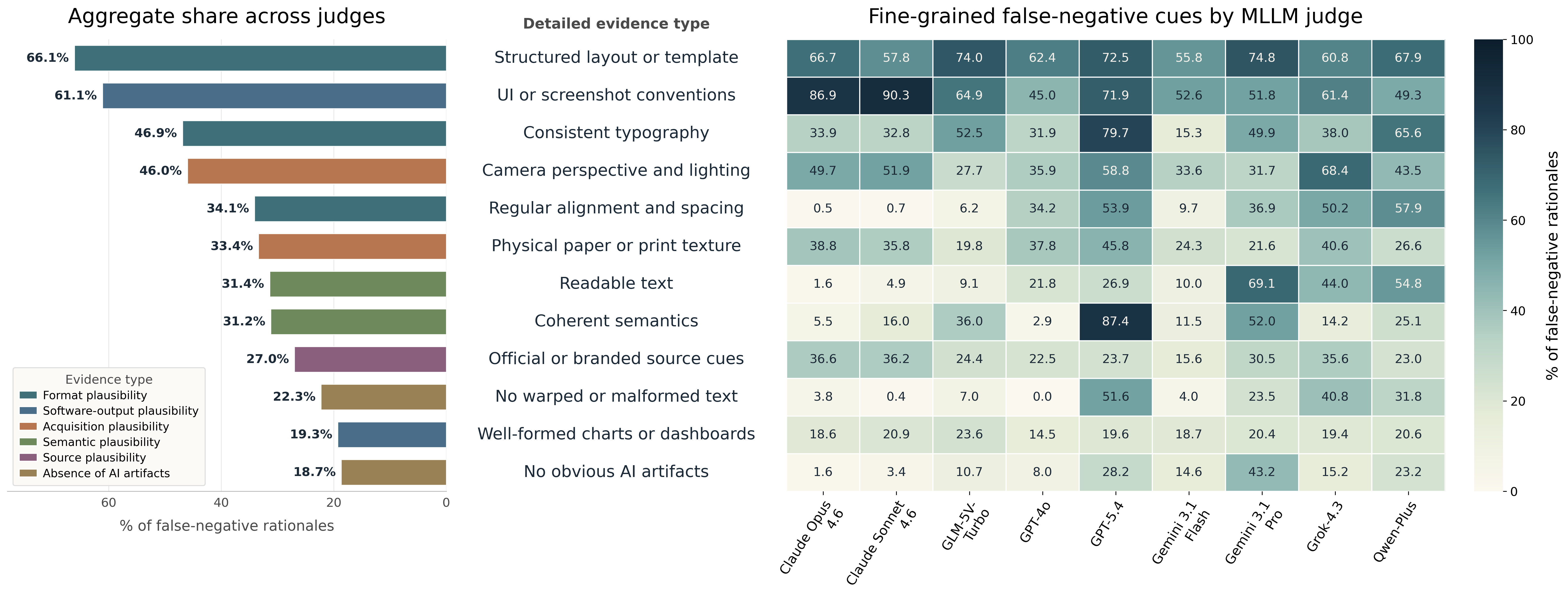}
  \vspace{-1.0ex}
  \caption{
  False-negative rationale cues for MLLM judges. Bars show aggregate non-exclusive cue frequencies, and the heatmap shows per-model distributions.
  }
  \label{fig:mllm_failure_reason}
  \vspace{-1ex}
\end{figure*}

\begin{figure}[!t]
  \centering
  \includegraphics[width=\linewidth]{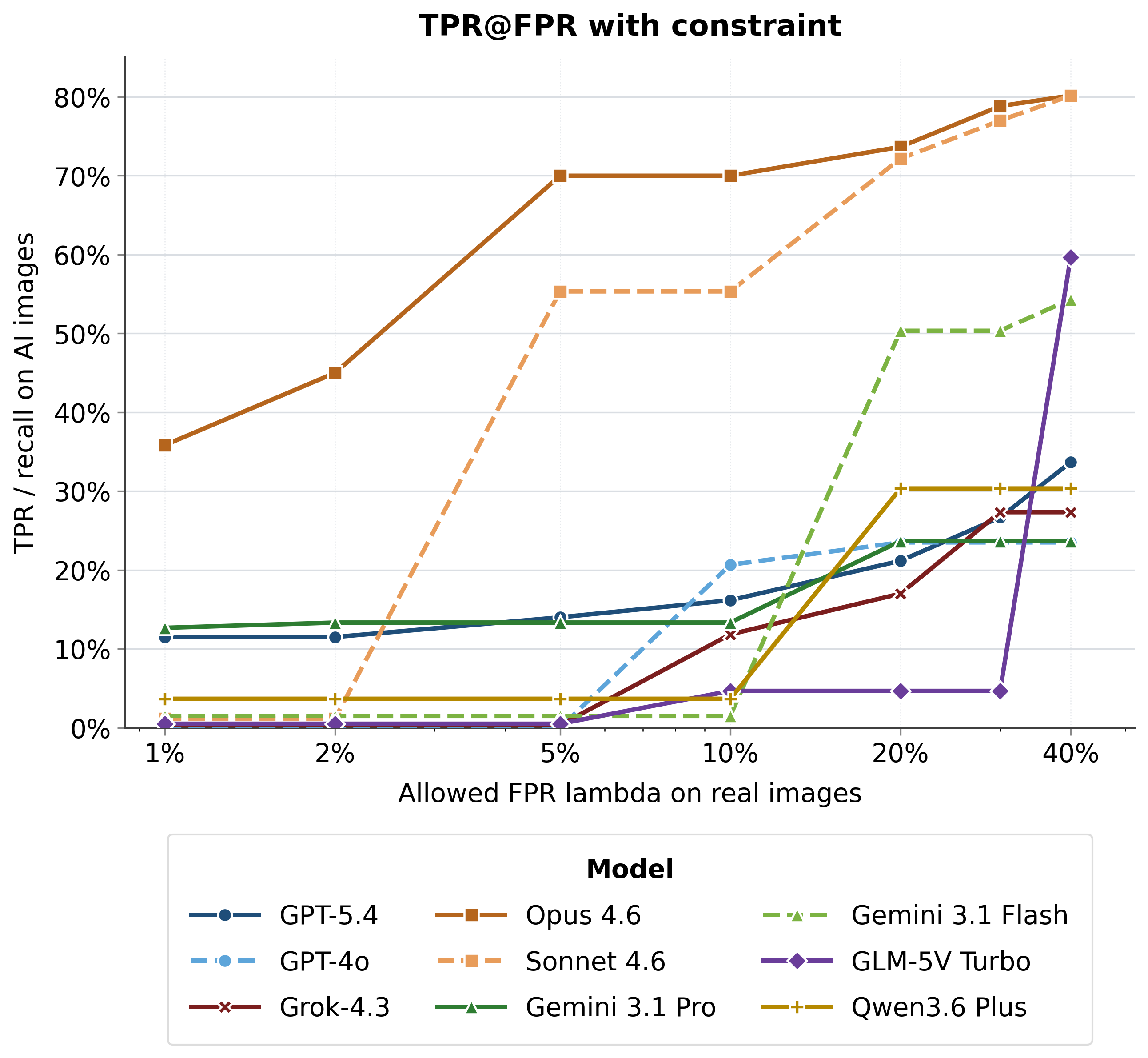}
  \vspace{-3.0ex}
  \caption{
  TPR results under increasing FPR budgets for MLLM judges.
  }
  \label{fig:tpr_fpr_analysis}
  \vspace{-2.2ex}
\end{figure}

\paragraph{MLLMs struggle to detect synthetic credibility.}
As shown in Table~\ref{tab:main-taxonomy-results}, MLLMs perform poorly on \textsc{SynCred-Bench}. Across all 15 MLLMs, the average positive-class accuracy is only 31.2\%, and the average TPR(5) is 10.5\%. Several strong closed-source models, including GPT-5.4, GPT-4o, Grok-4.3, and Gemini 3.1 Pro, obtain below 25\% accuracy, indicating that general visual reasoning ability does not directly translate to robust detection of credibility-bearing forgeries. Claude Opus 4.6 is the strongest usable MLLM, achieving 69.5\% accuracy with 5.0\% FPR. Performance is also uneven across settings: MLLMs are relatively better on credential records, but substantially weaker on screen-capture styles, where the average accuracy drops to 18.1\%.

\paragraph{Dedicated AIGC detectors are also not robust.} They perform better than MLLMs on average, but still fall short of reliable detection. Their average accuracy is 48.3\%, with an average TPR(5) of 30.3\%. Among them, Hive AI is the strongest calibrated detector, reaching 75.2\% accuracy with 0.9\% FPR. In contrast, AI-vs-Real obtains the highest raw accuracy of 78.7\%, but its 69.1\% FPR indicates poor calibration and many false alarms on real images. Other open-source detectors are much weaker, with AI-vs-Human and Deepfake-vs-Real achieving only 8.3\% and 19.3\% accuracy, respectively. These results suggest that existing AIGC detectors are not robust to synthetic credibility: they may detect some low-level generation traces, but they do not consistently handle document-like, interface-like, or circulation-style artifacts whose visual appearance differs from conventional natural-image deepfakes.

\begin{figure*}[!t]
  \centering
  \includegraphics[width=\linewidth]{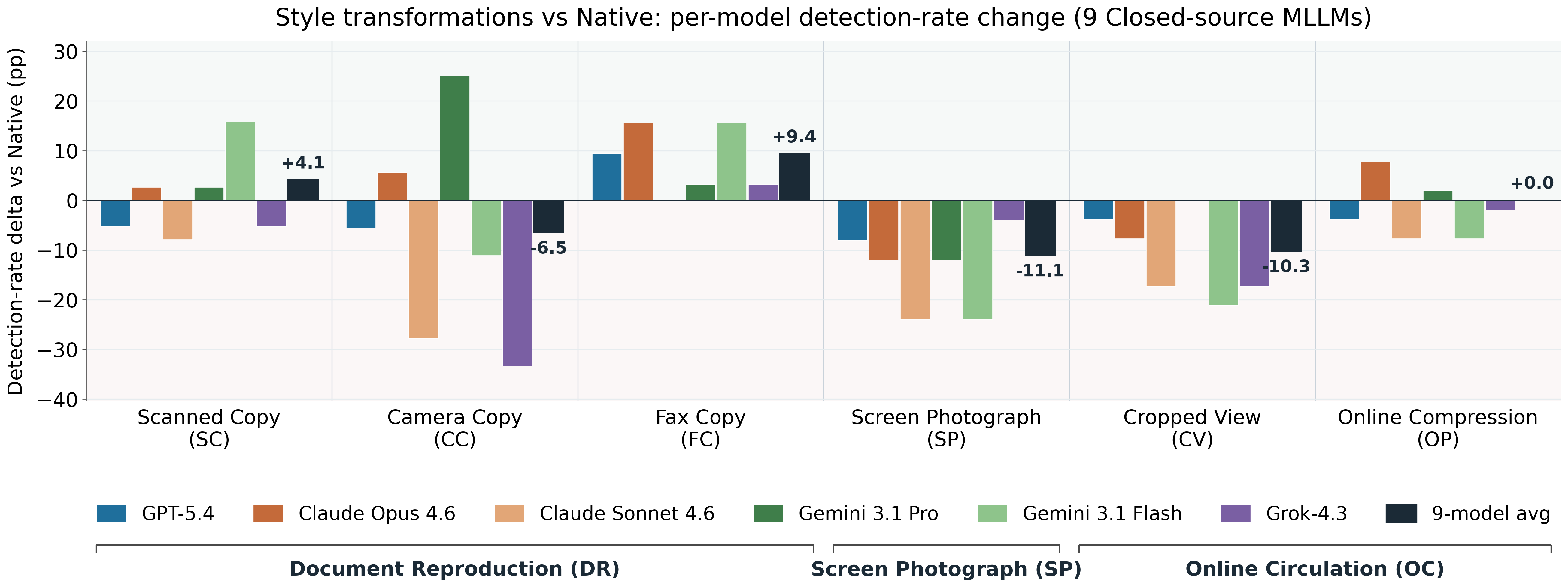}
  \vspace{-2.5ex}
  \caption{
  TPR change of each circulation style relative to Native Rendering for closed-source MLLM judges.
  }
  \label{fig:circulation_style_delta}
  \vspace{-1ex}
\end{figure*}

\subsection{Threshold sweeping reveals weak low-FPR operating points.}
\label{sec:tpr_fpr_analysis}

Figure~\ref{fig:tpr_fpr_analysis} evaluates selected MLLM judges by sweeping confidence thresholds on FP450. The main trend is that recall remains limited under strict false-positive budgets. At 5\% FPR, only Claude Opus 4.6 reaches a high TPR of about 69.5\% and Sonnet 4.6 reaches 55.3\%, while most other models stay below 30\%. Gemini 3.1 Flash, GLM-5V Turbo, and Qwen3.6 Plus improve mainly under looser FPR budgets, indicating that their detection signals are poorly separated from real-image scores.

\subsection{Credible layout is frequently converted into authenticity evidence.}
\label{sec:mllm_failure_reason}

The false-negative analysis in Figure~\ref{fig:mllm_failure_reason} shows that MLLM failures are driven by cues that make synthetic artifacts appear credible. Structured layouts/templates and UI/screenshot conventions dominate rationales, appearing in 66.1\% and 61.1\% of false-negative explanations; typography and camera perspective/lighting follow at 46.9\% and 46.0\%. These far exceed rationales based on the absence of visible AI artifacts. Per-model patterns differ: Claude Opus 4.6 and Claude Sonnet 4.6 emphasize UI/screenshot conventions, GPT-5.4 cites coherent semantics, typography, and layout, Grok-4.3 relies on camera perspective and lighting, and Qwen3.6 Plus highlights typography, alignment, and readable text. False negatives therefore are not merely missed synthesis errors; judges often reinterpret credible layout, coherent content, and plausible acquisition cues as authenticity evidence.

\subsection{Circulation styles can systematically distort authenticity judgments.}
\label{sec:circulation_style_delta}

Figure~\ref{fig:circulation_style_delta} shows that circulation style shifts MLLM detection in different directions relative to Native Rendering (NR). Camera Copy, Screen Capture, and Cropped View reduce detection by -6.5 pp, -11.1 pp, and -10.3 pp. These styles make the image more credible because blur, perspective distortion, partial content, and screenshot framing provide plausible explanations for visual defects, making them resemble ordinary capture or reposting rather than fabrication. Scanned Copy and Fax-like Copy increase average detection by +4.1 pp and +9.4 pp. Generic Degradation has almost no average effect, showing that semantically meaningful circulation cues matter more than compression alone. Overall, screen-like and repost-like styles weaken authenticity judgments by converting synthetic artifacts into plausible acquisition evidence.

\subsection{Humans Also Struggle to Distinguish Synthetic Credibility}

\begin{table}[!t]
\centering
\resizebox{\columnwidth}{!}{%
\begin{tabular}{lccc}
\toprule
Human Evaluator & TPR & FPR & Avg. Conf.\\
\midrule
Individual decisions & 60.2\% & 35.8\% & 3.64 / 5\\
Majority voting & 63.0\% & 27.0\% & -- \\
\bottomrule
\end{tabular}
}
\vspace{-0.2em}
\caption{Human annotation performance under individual decisions and majority vote.}
\vspace{-1.2em}
\label{tab:human_annotation_main}
\end{table}

To assess whether synthetic credibility poses a practical challenge for human judgment, we recruited 20 university students, including 4 Ph.D. students and 16 undergraduates from economics and finance, automation, computer science and technology, electronics, humanities and social sciences, and the arts. We constructed the annotation set by category-balanced sampling 100 AI-generated samples from the 600 samples in SynCred Bench and 100 samples from FP450, then mixing and randomly shuffling them into a unified test set. Each item was independently judged by five annotators. As shown in Table~\ref{tab:human_annotation_main}, human evaluators achieved only moderate performance: individual decisions reached a TPR of 60.2\% with a relatively high FPR of 35.8\%, while majority voting improved the TPR to 63.0\% and reduced the FPR to 27.0\%. These results suggest that synthetic credibility is difficult for humans to reliably identify and therefore represents a realistic and urgent risk for misinformation detection. Detailed results are listed in Appendix~\ref{app:human-eval}.

\subsection{How to Handle Synthetic Credibility}

(1) Provenance and watermark verification should be strengthened, since they provide direct evidence of image generation or editing and recent methods are increasingly robust to cropping, compression, and resizing~\cite{fernandez2023stable,wen2023treering,gowal2025synthidimage}.  
(2) AIGC detector training data should expand beyond natural scenes and portrait deepfakes to cover document-like, interface-like, and circulation-style forgeries with matched real negatives~\cite{zhu2023genimage}.  
(3) MLLMs should be aligned not to treat credible visual cues, such as seals, templates, typography, UI conventions, and compression artifacts, as evidence of authenticity, but instead separate visual plausibility from provenance and recommend external verification~\cite{sun2023llavarlhf,yu2024rlhfv}.

\section{Conclusion}

This paper identifies synthetic credibility as an emerging visual misinformation threat, where AI-generated artifacts become persuasive by imitating both authoritative formats and realistic circulation traces. To study this problem, we introduce \textsc{SynCred-Bench}, a benchmark of 600 AI-generated claim-bearing images spanning six credible-form categories and seven circulation styles, together with FP450 for evaluating false positives on real images. Extensive experiments show that current MLLMs and AIGC detectors remain unreliable, especially under low false-positive-rate constraints. Our analysis further reveals that models often mistake structured layouts, official-looking templates, typography, screenshots, lighting, and degradation artifacts for evidence of authenticity. And human annotators struggled to identify synthetic credibility, reaching only 63\% success. These findings suggest that synthetic credibility cannot be addressed by conventional natural-image forgery detection alone. Future systems should combine provenance verification, watermarking, broader detector training data, and MLLM alignment that explicitly separates visual plausibility from image provenance.

\section*{Limitations}

This work has several limitations. First, \textsc{SynCred-Bench} is intended as an initial benchmark for synthetic credibility rather than an exhaustive representation of all credible visual misinformation. Although we cover six credible-form categories and seven circulation styles, real-world misleading artifacts may involve additional formats, languages, cultural conventions, platform interfaces, and domain-specific credibility cues that are not included in the current dataset. The benchmark currently focuses on English and Chinese claim-bearing images, and its conclusions may not directly transfer to languages, scripts, regions, or institutional styles that are visually and socially different from those studied here.

Second, our synthetic set is generated with a single text-to-image generation pipeline, because gpt-image-2 is the only model capable of generating such samples at the moment. This may also introduce generator-specific artifacts, prompt-template regularities, or rendering biases. When more models are capable of generating such items, future work should extend the benchmark to multiple image-generation systems, editing-based pipelines, open-source generators, and adversarially post-processed samples. Such extensions are important because the capabilities and visual signatures of generative models change rapidly.

Third, our human verification procedure checks whether each sample contains the intended claim, matches the assigned credible-form category, and exhibits the intended circulation style. However, the labels are assigned by construction, and our validation does not constitute a large-scale perceptual study of whether ordinary users actually find each artifact credible. A user study with independent annotators would provide stronger evidence about the perceived credibility, persuasiveness, and potential real-world harm of different form--circulation combinations.

Fourth, due to limitations in computational resources and evaluation cost, our experiments are primarily conducted on 600 AI-generated images and 450 real-image false-positive samples. While this evaluation setting provides useful preliminary evidence, the relatively limited scale may introduce potential randomness in the reported results and may not fully rule out evaluation-specific overfitting or benchmarking artifacts. Future work should therefore consider constructing a larger and more diverse benchmark dataset to improve statistical stability, reduce the influence of sampling variation, and further mitigate the risk of evaluation hacking.

Fifth, our evaluation primarily measures AI-generated-image detection, not full veracity verification. In our setup, generated images are treated as the positive class and real images as the negative class. This does not determine whether the embedded claim is true, false, misleading, unsupported, or contextually misrepresented. In practice, a real image can contain misinformation, and a generated image can contain a true claim. Therefore, \textsc{SynCred-Bench} should be viewed as a benchmark for detecting credibility-bearing synthetic visual artifacts, not as a complete fact-checking or provenance-verification system.

Finally, model results may change over time. Several evaluated systems are closed-source or API-based, and their behavior can be affected by model updates, safety-policy changes, prompt formatting, image preprocessing, and threshold calibration. In addition, our rationale analysis relies on model-generated explanations, which are useful for identifying recurring error patterns but should not be interpreted as direct causal evidence of the internal decision process. Future work should complement rationale analysis with controlled interventions, counterfactual image edits, and more systematic attribution methods.

\section*{Ethical Considerations}

This work studies a dual-use problem: synthetic credibility-bearing artifacts such as AI-generated notices, screenshots, certificates, dashboards, and receipts can support detection research, but may also be misused for misinformation, impersonation, fraud, or deception. We therefore treat \textsc{SynCred-Bench} as a research resource for measurement and safety analysis, not as a tool for producing misleading content. All generated images should be clearly labeled as synthetic in documentation, metadata, and release materials. We recommend research-only access terms that prohibit deceptive use, political manipulation, harassment, impersonation, and public-facing misuse; dataset access should involve identity and intent verification, and any released prompts or metadata should include explicit restrictions and safety warnings.

We also emphasize the risk of false positives. Incorrectly labeling authentic images as AI-generated can harm journalists, institutions, document holders, platform users, and ordinary individuals. Detector outputs should not be used as the sole basis for accusing a source of fabrication. Real-world use should combine model predictions with provenance signals, watermark or content-credential checks, source verification, human review, and claim-level fact-checking; low-confidence results should be treated only as signals for further investigation.

The real-image negative set raises copyright, attribution, privacy, and data-governance concerns. Releases should respect source licenses, avoid redistributing sensitive personal information, document image sources where possible, and provide removal mechanisms. The benchmark may also reflect cultural, linguistic, and institutional assumptions from its construction process. Future expansions should improve geographic, linguistic, and cultural coverage, especially for high-stakes domains such as public administration, education, healthcare, finance, and emergency communication.

\bibliography{anthology, custom}
\bibliographystyle{acl_natbib}

\appendix

\section{Dataset Details}
\label{app:more_details_dataset}

\subsection{Detailed Dataset composition.}

The positive split contains 600 AI-generated images. The dataset is balanced by credible form, with 100 images for each of the six artifact categories: Media Layout (ML), Institutional Notice (IN), Platform Interface (PI), Credential Record (CR), Analytical Display (AD), and Assessment Material (AM). Within each category, we include multiple subtypes to avoid reducing the benchmark to a small number of repeated templates. When an image could plausibly belong to multiple categories, we assign the category according to the dominant social function of the artifact. For example, a payment confirmation embedded in a mobile interface is assigned to Credential Record if its main function is transactional proof.

\paragraph{Circulation Distribution} The circulation axis describes how an artifact appears to have been accessed, reproduced, photographed, cropped, degraded, or reshared. This axis is not forced to be exactly uniform across all form--style cells. Instead, we stratify the dataset to cover all circulation styles while preserving visually plausible form--style pairings. For example, scanned or photographed copies are natural for certificates, notices, and assessment materials, while cropped or compressed views are more natural for platform interfaces, media layouts, and online circulation. Table~\ref{tab:style-sample-counts} reports the final fine-grained circulation distribution after resolving duplicated identifiers. Native Rendering (NR) contains clean digital renderings without secondary reproduction cues. Reproduced Style (RS) groups Scanned Copy (SC), Camera Copy (CC), and Fax-like Copy (FC). Screen Photograph (SP) covers images presented as photos of screens. Circulated Content (CI) groups Cropped View (CV) and Online Compression (OC), corresponding to partial reposts and compression-degraded online images.

\begin{table}[t]
\centering
\renewcommand{\arraystretch}{1.15}
\resizebox{\columnwidth}{!}{%
\begin{tabular}{llrr}
\hline
\textbf{Style Category} & \textbf{Subtype} & \textbf{Abbrev.} & \textbf{Count} \\
\hline
Native Rendering & -- & NR & 152 \\
Document Reproduction & Scanned Copy & SC & 85 \\
Document Reproduction & Camera Copy & CC & 104 \\
Document Reproduction & Fax Copy & FC & 49 \\
Screen Photograph & -- & SP & 64 \\
Online Circulation & Cropped View & CV & 65 \\
Online Circulation & Online Compression & OP & 81 \\
\hline
\multicolumn{3}{r}{\textbf{Total}} & 600 \\
\hline
\end{tabular}%
}
\caption{Sample counts by presentation style after removing the duplicated ID sample. Reproduced Style (RS) is the union of SC, CC, and FC; Circulated Content (CI) is the union of CV and OC.}
\label{tab:style-sample-counts}
\end{table}


\paragraph{Languages and domains} The generated images contain misinformation claims in both Chinese and English. The final dataset contains 304 English claim source and 296 Chinese claim source. The prompts and metadata record the language of the rendered content. The semantic topics span education, public administration, finance, healthcare, commerce, online platforms, rankings, and analytical reports. This coverage is intended to test whether detectors can handle credibility-bearing artifacts across different institutional and everyday information formats.

\subsection{FP450 Negative Set Details}
\label{app:fp_detail}

We construct FP450 as a matched real-image negative set for false-positive evaluation. FP450 is collected from web image search using queries that cover the same broad visual space as SYNCRED-BENCH, including notices, screenshots, certificates, receipts, invoices, dashboards, scanned documents, photographed documents, cropped views and compressed online images. One of the key challenges in constructing our dataset is the lack of diverse, stylistically balanced image corpora on the public internet. Specifically, styles such as Screen Captures, Camera copies and Cropped views as well as compressed styles are relatively underrepresented due to preferences of clear, full-scale images for both content providers and platforms. Moreover, collecting and reproducing large-scale real-world document artifacts would introduce practical constraints, including copyright concerns, privacy considerations and implicit bias due to cultural and geographical contexts. 

Therefore, after collecting diverse and non-duplicate raw images from web queries, we manually edited images to emulate specific and consistent styles. For Fax-like Copies and Cropped Views, we used built-in image editing tools to simulate styles. Specifically, Fax-like Copies are generated by reducing sharpness and saturation until obtaining monochrome images, while Cropped views are the result of truncation. For Screen Captures, we directly  photograph displayed images on physical screens to preserve natural screen reflections and Moiré patterns. For Degradations, we downscale images and recapture them through screenshots to imitate the compressed and blunt features of degraded artifacts. When available, we inspect source-page context to check whether the image appears in an ordinary real image setting, such as an official page, product support page, news article, documentation page, platform help page, or user-facing interface capture. 

Data distribution in FP450 is listed in Table~\ref{tab:fp-set-stats-fine-circulation}:

\begin{table}[!t]
\centering
\renewcommand{\arraystretch}{1.18}
\resizebox{\columnwidth}{!}{
\begin{tabular}{l|rrrrrrr|rr}
\hline
\multirow{2}{*}{\textbf{Credible Form}} &
\multicolumn{7}{c|}{\textbf{Fine-Grained Credible Circulation}} &
\multirow{2}{*}{\textbf{Total}} &
\multirow{2}{*}{\textbf{Share (\%)}} \\
\cline{2-8}
 & \textbf{NR} & \textbf{FC} & \textbf{CC} & \textbf{SC} & \textbf{SP} & \textbf{OP} & \textbf{CV} & & \\
\hline
\textbf{ML} & 11 & 13 & 7 & 13 & 11 & 9 & 6 & 70 & 15.6 \\
\textbf{IN} & 11 & 16 & 4 & 15 & 12 & 10 & 11 & 79 & 17.6 \\
\textbf{PI} & 12 & 7 & 6 & 13 & 10 & 9 & 10 & 67 & 14.9 \\
\textbf{CR} & 10 & 12 & 3 & 14 & 18 & 11 & 10 & 78 & 17.3 \\
\textbf{AD} & 14 & 14 & 13 & 14 & 12 & 14 & 14 & 95 & 21.1 \\
\textbf{AM} & 9 & 7 & 13 & 13 & 3 & 9 & 7 & 61 & 13.6 \\
\hline
\textbf{Total} & 67 & 69 & 46 & 82 & 66 & 62 & 58 & 450 & 100.0 \\
\textbf{Share (\%)} & 14.9 & 15.3 & 10.2 & 18.2 & 14.7 & 13.8 & 12.9 & 100.0 & -- \\
\hline
\end{tabular}
}

\vspace{-0.2em}
\caption{Fine-grained statistics of the FP real-image set across credible form and credible circulation.}
\vspace{-0.6em}
\label{tab:fp-set-stats-fine-circulation}
\end{table}

\subsection{Metadata Format}
\label{app:metadata-format}

Each generated image is paired with one metadata record. The record stores the image filename, semantic category labels, the exact generation prompt, and a compact metadata block for grouping and audit. During evaluation, detectors are shown only the image. The metadata is used only for dataset
construction, stratified reporting, and error analysis.

\begin{table*}[!t]
\centering
\small
\setlength{\tabcolsep}{5pt}
\renewcommand{\arraystretch}{1.12}
\begin{tabularx}{0.96\textwidth}{@{}L{0.25\textwidth}L{0.10\textwidth}Y@{}}
\toprule
\textbf{Field} & \textbf{Type} & \textbf{Meaning} \\
\midrule
\texttt{id}
& integer
& Stable item identifier within a category. \\

\texttt{image}
& string
& Image filename, usually following the pattern \texttt{PREFIX\_STYLE\_ID.ext}. \\

\texttt{labels.artifact\_type}
& string
& High-level artifact category. \\

\texttt{labels.generation\_method}
& string
& Generation route. Most samples use \texttt{direct}; a small number of media artifacts use \texttt{edit}. \\

\texttt{labels.style}
& string
& Presentation or acquisition style label. Raw style-name typos are normalized before analysis. \\

\texttt{prompt}
& string
& Full image-generation prompt, including both semantic content and style-control cues. \\

\texttt{metadata.subtype}
& string
& Fine-grained artifact subtype, such as dashboard, receipt, transcript, or social-media page. \\

\texttt{metadata.subject}
& string
& Semantic topic or target content of the artifact. \\

\texttt{metadata.lang}
& string
& Language tag for the prompt or rendered content, mainly \texttt{zh} or \texttt{en}. \\
\bottomrule
\end{tabularx}
\caption{Metadata schema used by \texttt{index.json} and category-level \texttt{meta.json} files.}
\label{tab:metadata-schema}
\end{table*}

\noindent
Table~\ref{tab:metadata-schema} summarizes the stored fields. A representative metadata record, translated to English for readability, is shown in Figure~\ref{fig:metadata-record-example}. The language tag reflects the original prompt or rendered content.

\begin{figure*}[t]
\centering
\begin{minipage}{0.96\textwidth}
\begin{metadatajson}
{
  "id": 0,
  "image": "RP_Native_0.png",
  "labels": {
    "artifact_type": "Analytical Display",
    "generation_method": "direct",
    "style": "Native"
  },
  "prompt": "Create a BI dashboard exported from Metabase/Superset about monthly new-energy vehicle sales trends, with KPI cards, trend charts, category shares, filters, and ranking tables. Render it as a high-resolution native digital export with sharp fonts and clean chart boundaries, with no scanning, camera, or repost artifacts.",
  "metadata": {
    "subtype": "dashboards",
    "subject": "monthly new-energy vehicle sales trends",
    "lang": "zh"
  }
}
\end{metadatajson}
\end{minipage}
\caption{Example metadata record. The prompt text is translated to English for readability.}
\label{fig:metadata-record-example}
\end{figure*}

\subsection{Data Release and Access Protocol.}
\label{app:release}
Because SYNCRED-BENCH contains dual-use credibility-bearing artifacts, we adopt a controlled release protocol rather than unrestricted public redistribution. We will publicly release the benchmark documentation, category and circulation taxonomies, metadata schema, fixed evaluation splits, evaluation code, aggregate statistics, and model prediction files. The full image set, generation prompts, and prompt-level metadata will be distributed only for research use under a data-use agreement that prohibits deceptive use, impersonation, public redistribution, political manipulation, fraud, harassment, and training systems to generate misleading documents. Each released sample will be accompanied by explicit synthetic-origin labels, version identifiers, and cryptographic hashes to support reproducibility and auditing. For FP450, we will release only images with compatible licenses or provide source references, derived metadata, and hashes when redistribution is not permitted.

\section{MLLM Judge Details}
\label{app:mllm-judge-details}

We evaluate multimodal large language models (MLLMs) as image-level detectors. Each model receives only the image and a fixed detection instruction. No metadata fields, category labels, generation prompts, or style labels are provided to the detector.

We use a unified prompt to instruct all models to output the final judgment, confidence, and the reasons behind the analysis. The specific prompt is as follows:

\begin{tcolorbox}[title = {Prompt Template of GPT-based Evaluation}, breakable]
You are an AI-generated image detector. Judge whether the submitted image was generated by AI. Return a compact machine-readable response with exactly three fields:

\{
    
  "ai\_generated": true or false.
  
  "confidence": a number from 0.0 to 1.0.
  
  "reason": 2--4 concise sentences explaining the visual evidence for the judgment.

\}
\end{tcolorbox}

\section{Details of Human Detection Experiment}
\label{app:human-eval}
We recruited 20 university students who were unaware of the data construction pipeline used in this study to perform the annotation task. The annotators included 4 Ph.D. students and 16 undergraduate students, with academic backgrounds spanning economics and finance, automation, computer science and technology, electronics, humanities and social sciences, and the arts.

From the 600 AI-generated samples in SynCred-Bench, we selected 100 samples through category-balanced sampling. We also selected 100 samples from FP450 using the same balanced sampling strategy. The two sets were then combined and randomly shuffled to form the annotation questions. Each question was independently evaluated by five participants, who provided their judgment labels.

The results show that human annotators achieved an average accuracy of 63\% in identifying synthetic credibility, with the best individual annotator achieving no more than 80\% overall accuracy. The annotations also exhibited a relatively high false positive rate. These findings demonstrate that the phenomenon investigated in this paper has strong real-world significance and urgency, as it poses a genuine risk to society’s ability to distinguish false or misleading information.

Detailed annotator-level statistics, with all private information removed, are reported in Table~\ref{tab:human_annotation_subjects}.

\begin{table}[!t]
\centering
\resizebox{\columnwidth}{!}{%
\begin{tabular}{clccc}
\toprule
Subject & Subset & Overall accuracy & TPR & FPR \\
\midrule
1 & subset\_01 & 54.0\% & 76.0\% & 68.0\% \\
2 & subset\_01 & 54.0\% & 36.0\% & 28.0\% \\
3 & subset\_01 & 58.0\% & 64.0\% & 48.0\% \\
4 & subset\_01 & 64.0\% & 68.0\% & 40.0\% \\
5 & subset\_01 & 66.0\% & 80.0\% & 48.0\% \\
6 & subset\_02 & 72.0\% & 88.0\% & 44.0\% \\
7 & subset\_02 & 56.0\% & 52.0\% & 40.0\% \\
8 & subset\_02 & 62.0\% & 48.0\% & 24.0\% \\
9 & subset\_02 & 60.0\% & 64.0\% & 44.0\% \\
10 & subset\_02 & 68.0\% & 48.0\% & 12.0\% \\
11 & subset\_03 & 62.0\% & 52.0\% & 28.0\% \\
12 & subset\_03 & 66.0\% & 56.0\% & 24.0\% \\
13 & subset\_03 & 66.0\% & 60.0\% & 28.0\% \\
14 & subset\_03 & 64.0\% & 40.0\% & 12.0\% \\
15 & subset\_03 & 68.0\% & 48.0\% & 12.0\% \\
16 & subset\_04 & 66.0\% & 64.0\% & 32.0\% \\
17 & subset\_04 & 44.0\% & 52.0\% & 64.0\% \\
18 & subset\_04 & 78.0\% & 68.0\% & 12.0\% \\
19 & subset\_04 & 58.0\% & 72.0\% & 56.0\% \\
20 & subset\_04 & 58.0\% & 68.0\% & 52.0\% \\
\bottomrule
\end{tabular}
}
\caption{Anonymous subject-level human annotation performance.}
\label{tab:human_annotation_subjects}
\end{table}

\section{Examples of MLLM Failure}
\label{sec:appendix_mllm_failure_examples}

This section provides representative false-negative cases produced by multimodal large language model (MLLM) judges in our AI-generated image detection task. In each example, the input image is AI-generated, but the corresponding MLLM judge incorrectly classifies it as real with high confidence. We include the judge identity, confidence score, and the original free-form rationale returned by the model. These examples illustrate several recurring failure modes: MLLMs often over-rely on visually plausible layouts, coherent typography, realistic camera artifacts, and familiar interface conventions when assessing image authenticity. As a result, synthetic screenshots, document-like images, presentation photos, and broadcast-style scenes can be mistaken for genuine real-world captures when they contain consistent text, recognizable UI elements, or naturalistic lighting and perspective cues.

\begin{figure*}[!t]
    \centering
    \includegraphics[width=1\linewidth]{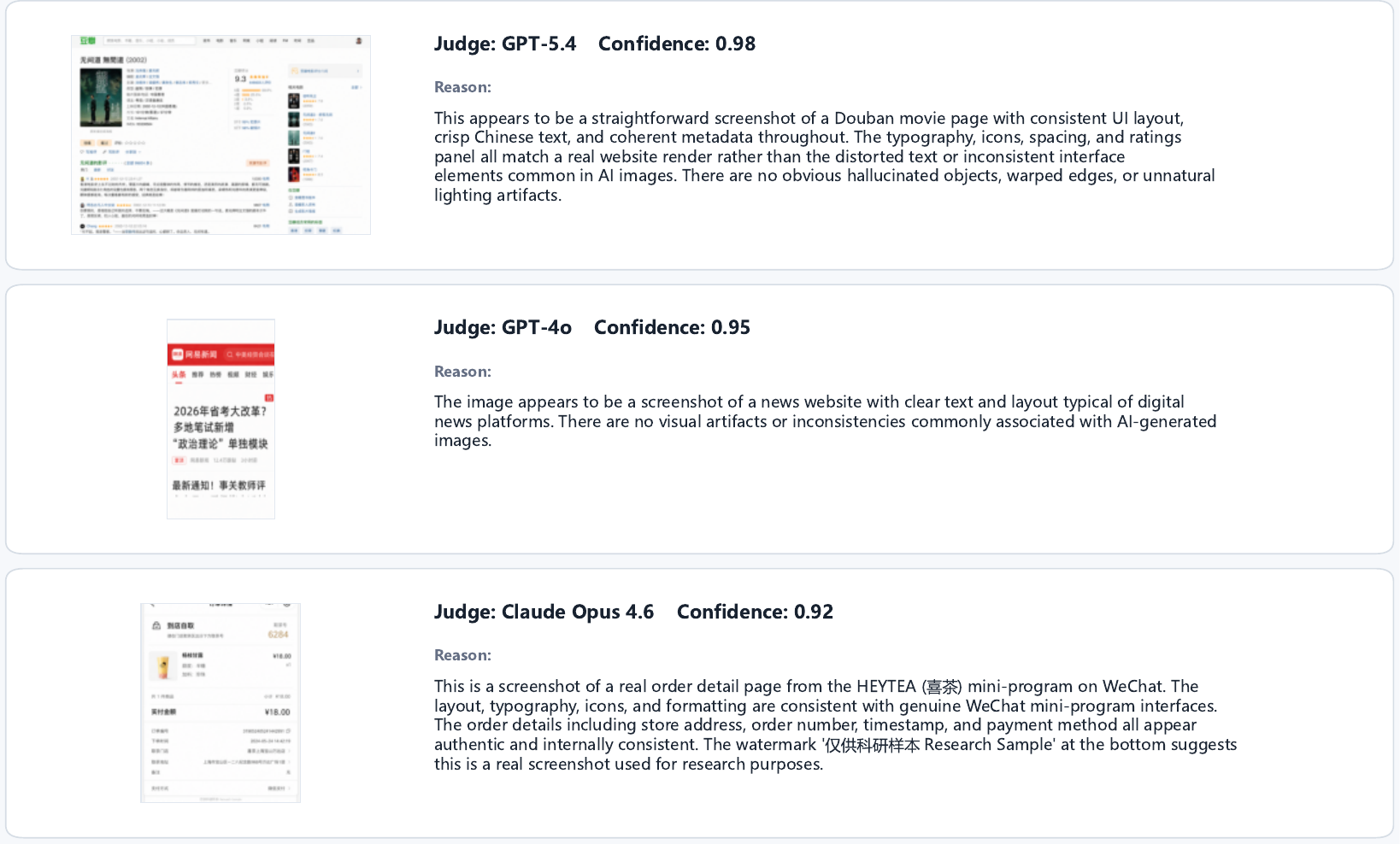}
    \caption{
    Representative false-negative examples from MLLM judges, Part 1.
    All displayed images are AI-generated, yet the judges classify them as real
    with high confidence. The examples include synthetic screenshots resembling
    a Douban movie page, a news website, and a HEYTEA order-detail interface.
    The model rationales show that the judges rely heavily on apparently
    consistent UI layout, crisp text rendering, coherent metadata, and the
    absence of obvious visual artifacts.
    }
    \label{fig:example_mllm_1}
\end{figure*}

\begin{figure*}[!t]
    \centering
    \includegraphics[width=1\linewidth]{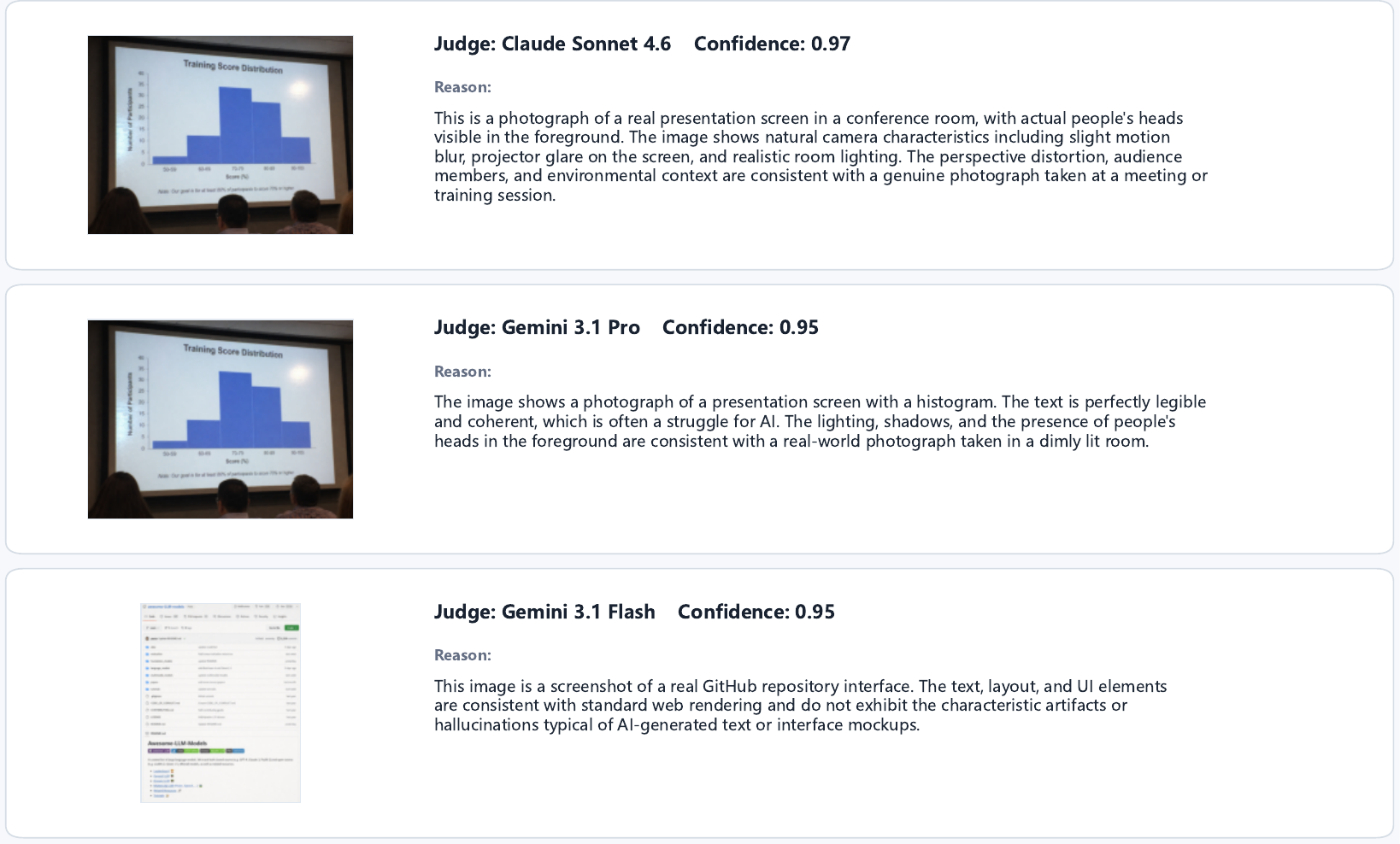}
    \caption{
    Representative false-negative examples from MLLM judges, Part 2.
    The shown AI-generated images are incorrectly identified as real by the
    judges. Two examples depict a presentation screen in a conference-room-like
    setting, while another resembles a GitHub repository interface. The
    rationales indicate that the judges mistake plausible environmental cues,
    such as projector glare, audience silhouettes, dim room lighting, legible
    text, and standard web-interface structure, as evidence of real image
    provenance.
    }
    \label{fig:example_mllm_2}
\end{figure*}

\begin{figure*}[!t]
    \centering
    \includegraphics[width=1\linewidth]{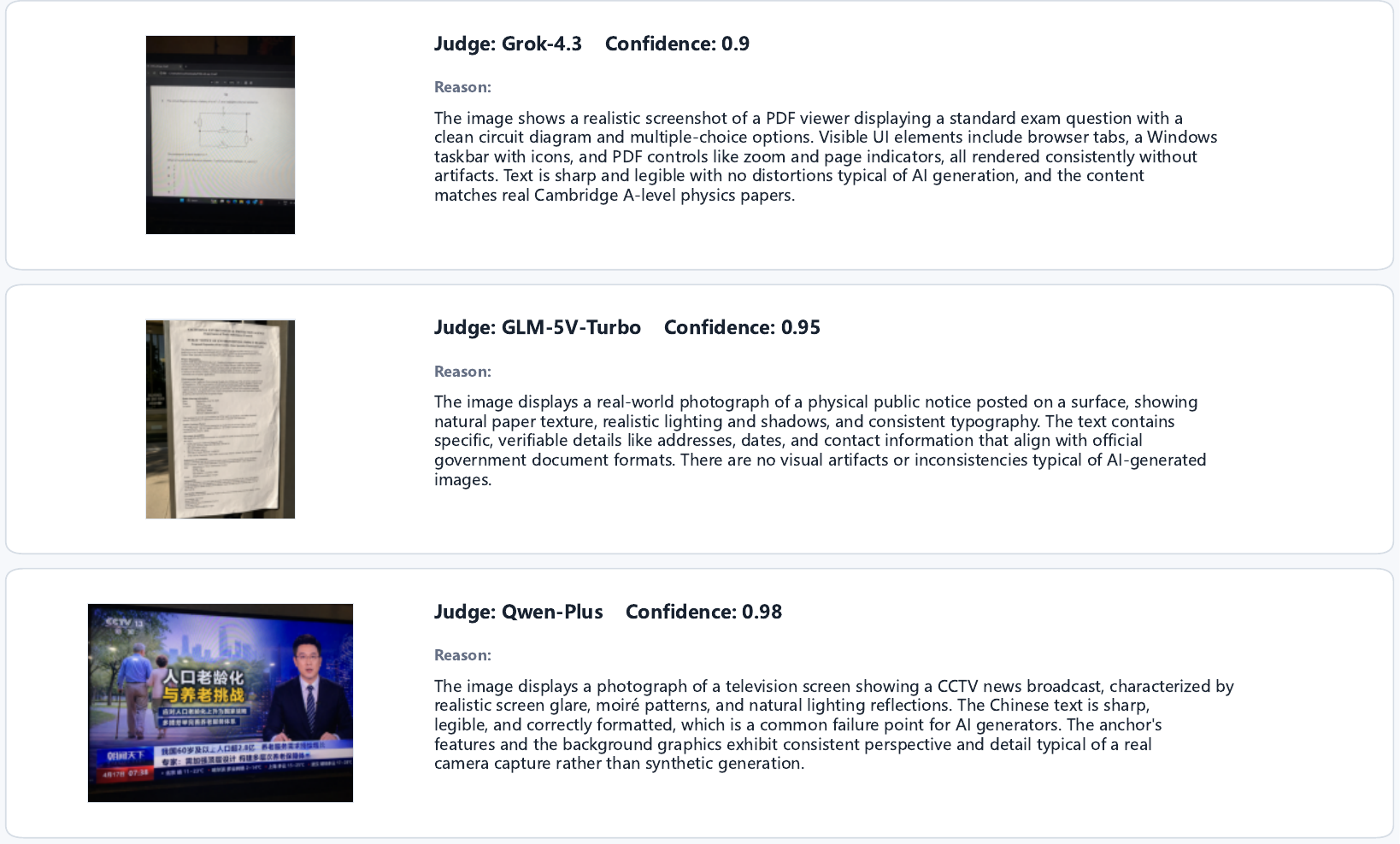}
    \caption{
    Representative false-negative examples from MLLM judges, Part 3.
    These AI-generated images are misclassified as real despite depicting
    synthetic document, notice, and television-broadcast scenes. The judges'
    explanations emphasize sharp text, familiar document formats, realistic
    paper texture, screen glare, moir\'e patterns, and coherent broadcast
    graphics. These cases suggest that MLLMs may conflate surface-level
    realism and recognizable visual conventions with genuine photographic
    authenticity.
    }
    \label{fig:example_mllm_3}
\end{figure*}

\section{Models Used in Our Experiments}
\label{app:models-used}

\subsection{MLLMs}

\paragraph{(1) Closed-source models.}
\begin{packeditemize}
    \item GPT-5.4: \url{https://developers.openai.com/api/docs/models/gpt-5.4}
    \item GPT-4o: \url{https://developers.openai.com/api/docs/models/gpt-4o}
    \item Grok-4.3: \url{https://docs.x.ai/developers/models/grok-4.3}
    \item Claude Opus 4.6: \url{https://www.anthropic.com/claude/opus}
    \item Claude Sonnet 4.6: \url{https://www.anthropic.com/claude/sonnet}
    \item Gemini 3.1 Pro: \url{https://deepmind.google/models/model-cards/gemini-3-1-pro/}
    \item Gemini 3.1 Flash-Lite: \url{https://deepmind.google/models/model-cards/gemini-3-1-flash-lite/}
    \item GLM-5V Turbo: \url{https://docs.z.ai/guides/vlm/glm-5v-turbo}
    \item Qwen3.6 Plus: \url{https://www.alibabacloud.com/help/en/model-studio/text-generation-model}
\end{packeditemize}

\paragraph{(2) Open-source models.}
\begin{packeditemize}
    \item Qwen3.5-9B: \url{https://huggingface.co/Qwen/Qwen3.5-9B}
    \item Qwen3.5-35B-A3B: \url{https://huggingface.co/Qwen/Qwen3.5-35B-A3B}
    \item Qwen3-VL-8B: \url{https://huggingface.co/Qwen/Qwen3-VL-8B-Instruct}
    \item Qwen3-VL-32B: \url{https://huggingface.co/Qwen/Qwen3-VL-32B-Instruct}
    \item Pixtral Large: \url{https://huggingface.co/mistralai/Pixtral-Large-Instruct-2411}
    \item Llama-3.2-11B-Vision: \url{https://huggingface.co/meta-llama/Llama-3.2-11B-Vision-Instruct}
\end{packeditemize}

\subsection{AIGC Detectors}

\paragraph{(1) Closed-source detectors.}
\begin{packeditemize}
    \item Sightengine: \url{https://sightengine.com/docs/ai-generated-image-detection}
    \item AI or Not: \url{https://docs.aiornot.com/api-reference/reports-by-modality/image}
    \item Hive AI: \url{https://docs.thehive.ai/docs/ai-image-and-video-detection}
\end{packeditemize}

\paragraph{(2) Open-source detectors.}
\begin{packeditemize}
    \item AI-vs-Real: \url{https://huggingface.co/dima806/ai_vs_real_image_detection}
    \item AI-vs-Human: \url{https://huggingface.co/dima806/ai_vs_human_generated_image_detection}
    \item Deepfake-vs-Real: \url{https://huggingface.co/dima806/deepfake_vs_real_image_detection}
\end{packeditemize}

\section{Data Examples}

\begin{figure*}[t]
\centering
\includegraphics[width=\textwidth,page=1]{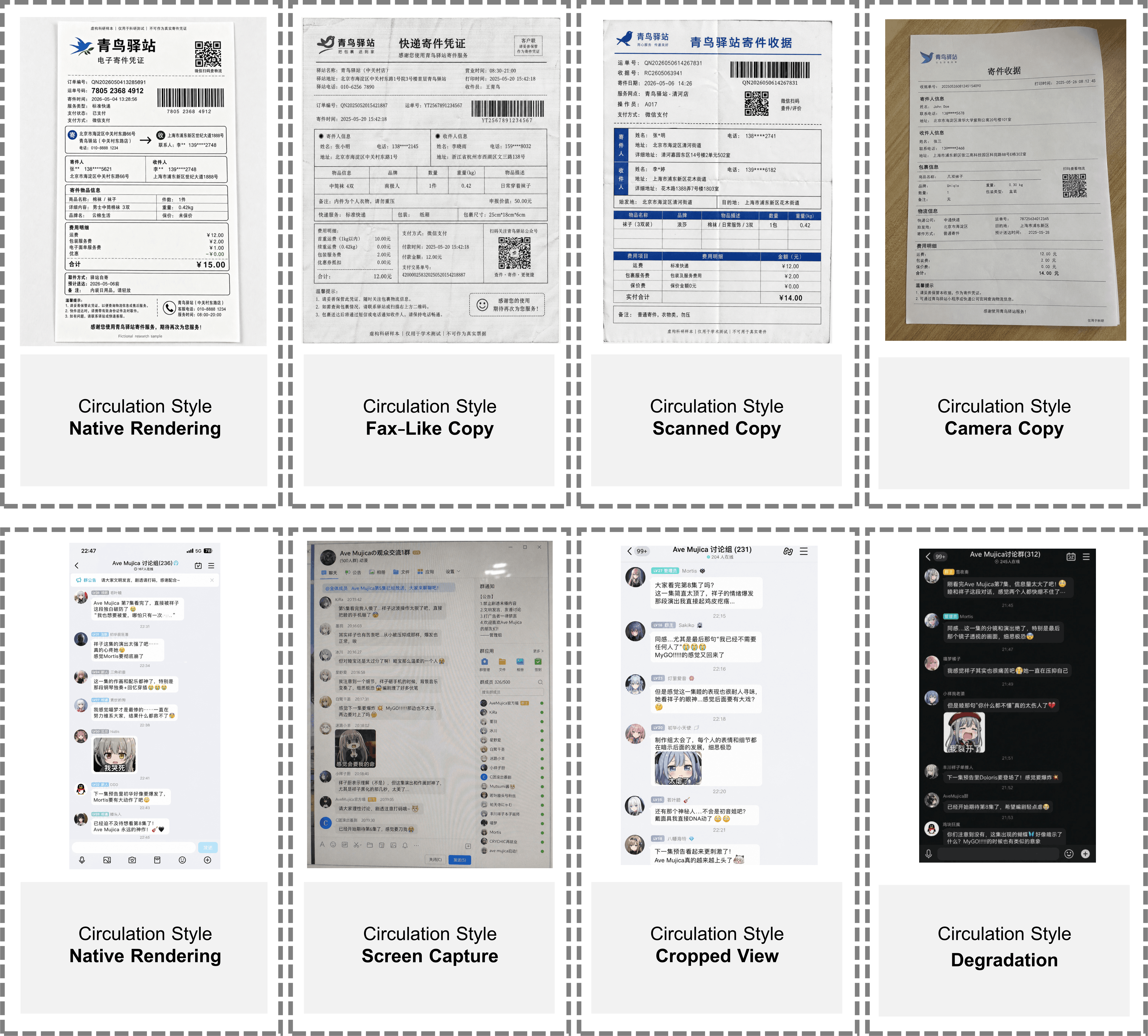}
\caption{
Examples by circulation style. Some images offer different circulation style variants for the same content.
}
\label{tab:data_examples_artifact_type}
\end{figure*}

\begin{figure*}[t]
\centering
\includegraphics[width=\textwidth,page=2]{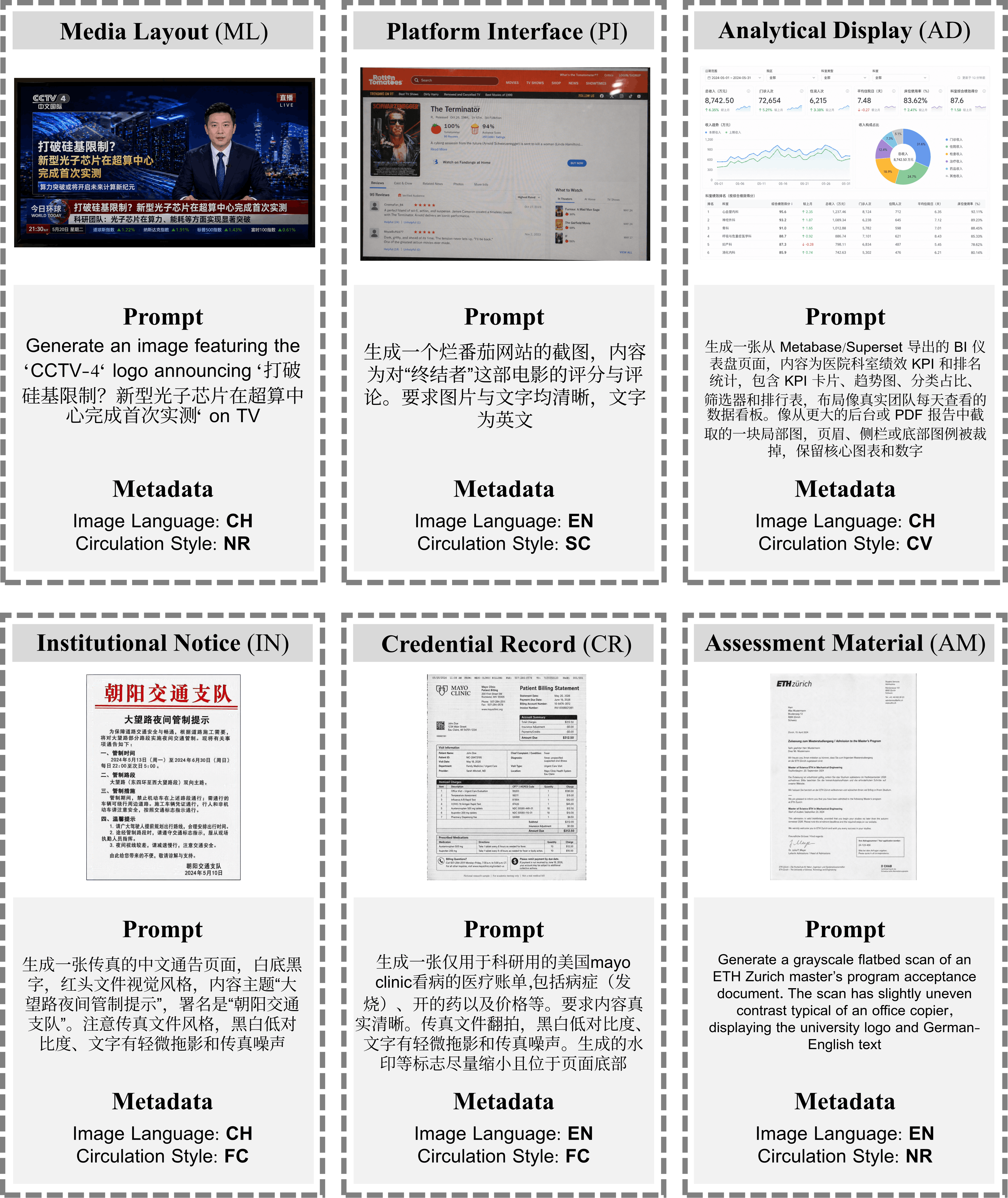}
\caption{
Examples by artifact type. Each row shows one randomly sampled example from an artifact type.
The circulation style is not fixed and is reported in the metadata field.
}
\label{tab:data_examples_circulation_style}
\end{figure*}

\begin{figure*}[t]
\centering
\includegraphics[width=\textwidth,page=1]{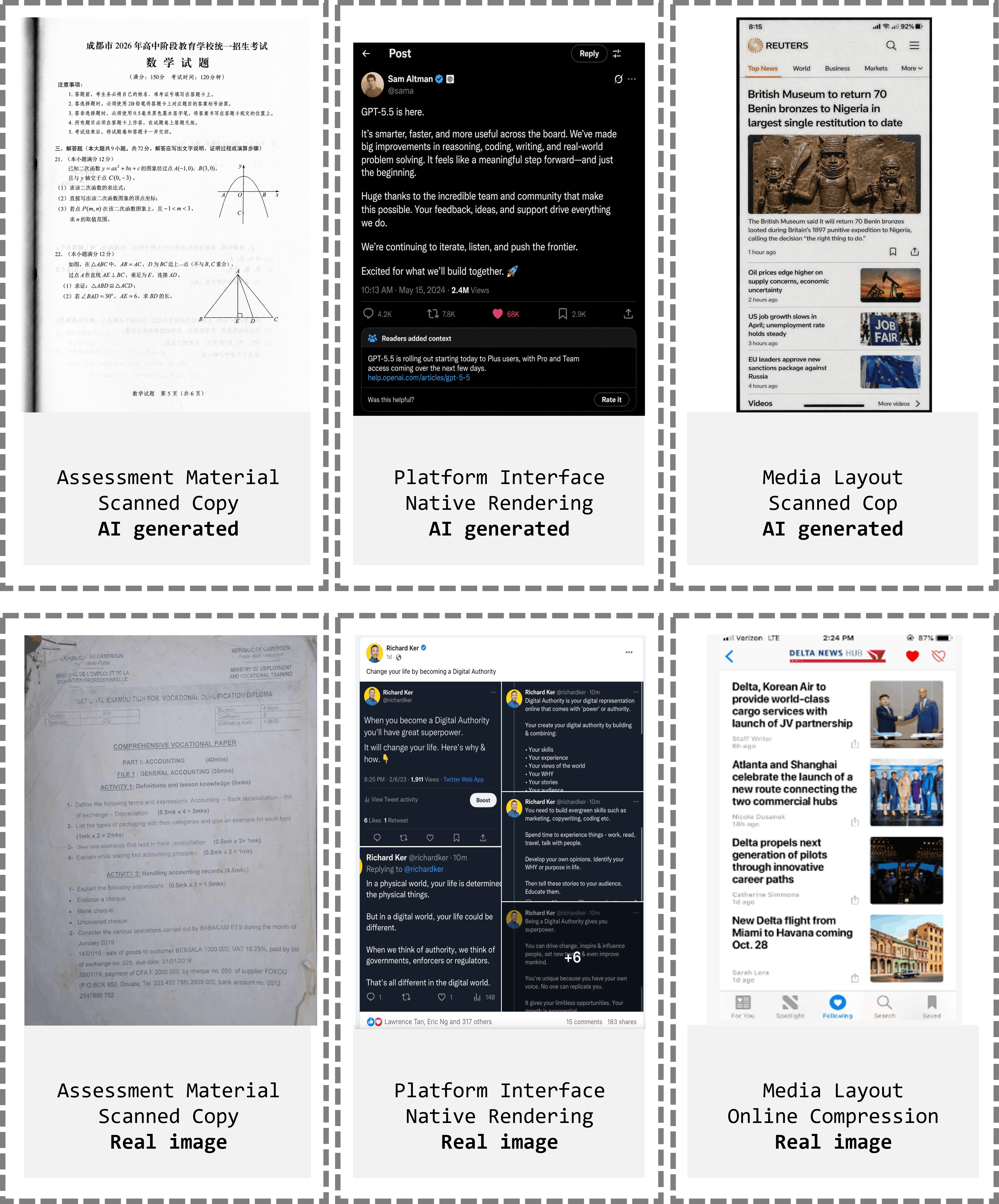}
\caption{
Examples of document images grouped by artifact type and image provenance. The top row shows AI-generated samples and the bottom row shows real images, covering assessment materials, platform-interface native renderings, and media-layout pages. These examples illustrate visual variations across scanned copies, native digital renderings, and compressed online layouts.
}
\label{tab:data_examples_artifact_type2}
\end{figure*}

\begin{figure*}[t]
\centering
\includegraphics[width=\textwidth,page=2]{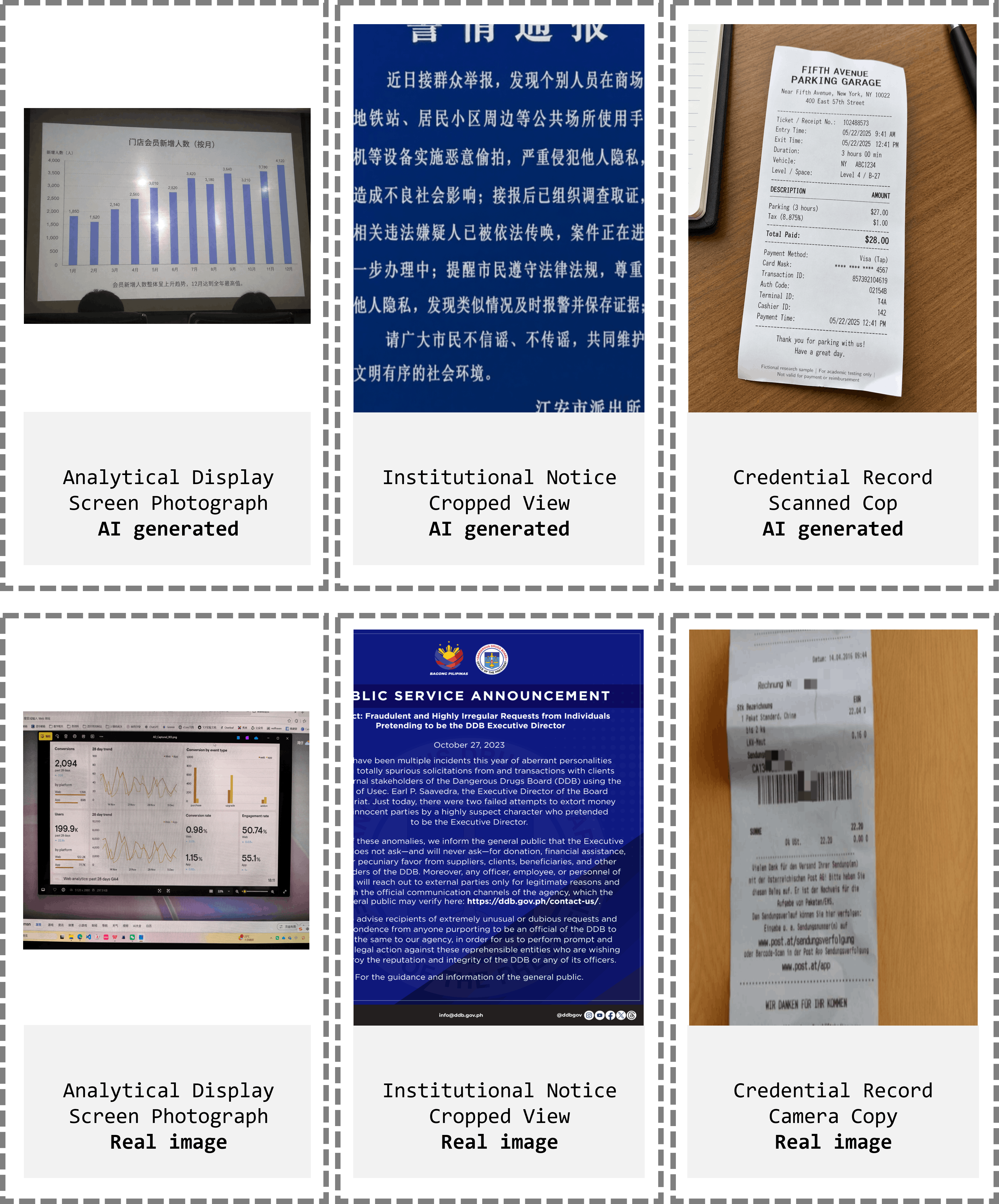}
\caption{
Examples of document images grouped by circulation and capture style. The top row shows AI-generated samples and the bottom row shows real images, including screen photographs of analytical displays, cropped views of institutional notices, and credential or receipt records captured as scanned or camera copies. These examples highlight realistic acquisition artifacts such as screen glare, cropping, perspective distortion, blur, and physical-document degradation.
}
\label{tab:data_examples_circulation_style2}
\end{figure*}

We provide additional examples of different styles and forms in Figure~\ref{tab:data_examples_artifact_type} and Figure~\ref{tab:data_examples_circulation_style}, and present several sets of comparative results between real images and AI-generated images in Figure~\ref{tab:data_examples_artifact_type2} and Figure~\ref{tab:data_examples_circulation_style2}.

\end{document}